\documentclass[journal]{IEEEtran}

\usepackage{comment}
\usepackage[ruled,vlined]{algorithm2e}
\usepackage{url}
\usepackage{textcomp,gensymb}
\usepackage{cite}
\usepackage{amsmath,amssymb,amsfonts}
\usepackage{graphicx}
\usepackage{textcomp}
\usepackage{xcolor}
\usepackage{multirow}
\usepackage{booktabs}
\usepackage{hyperref}

\usepackage{xcolor}

\usepackage{xcolor}

\usepackage{graphicx}

\begin{document}

\title{\LARGE \bf MagicGripper: A Multimodal Sensor-Integrated Gripper for Contact-Rich Robotic Manipulation}

\author{Wen Fan*, Haoran Li*, Dandan Zhang
\thanks{W. F and D. Zhang are with the Department of Bioengineering, Imperial College London. H. Li is with the School of Robotics, Xi'an Jiaotong-Liverpool University. \textit{Wen Fan and Haoran Li contributed equally to this work.}}
}

\maketitle

\begin{abstract}
Contact-rich manipulation in unstructured environments demands precise, multimodal perception to enable robust and adaptive control. Vision-based tactile sensors (VBTSs) have emerged as an effective solution; however, conventional VBTSs often face challenges in achieving compact, multi-modal functionality due to hardware constraints and algorithmic complexity. In this work, we present MagicGripper, a multimodal sensor-integrated gripper designed for contact-rich robotic manipulation. Building on our prior design, MagicTac, we develop a compact variant, mini-MagicTac, which features a three-dimensional, multi-layered grid embedded in a soft elastomer. MagicGripper integrates mini-MagicTac, enabling high-resolution tactile feedback alongside proximity and visual sensing within a compact, gripper-compatible form factor. We conduct a thorough evaluation of mini-MagicTac’s performance, demonstrating its capabilities in spatial resolution, contact localization, and force regression. We also assess its robustness across manufacturing variability, mechanical deformation, and sensing performance under real-world conditions. Furthermore, we validate the effectiveness of MagicGripper through three representative robotic tasks: a teleoperated assembly task, a contact-based alignment task, and an autonomous robotic grasping task. Across these experiments, MagicGripper exhibits reliable multimodal perception, accurate force estimation, and high adaptability to challenging manipulation scenarios.
Our results highlight the potential of MagicGripper as a practical and versatile tool for embodied intelligence in complex, contact-rich environments.


\end{abstract}



\IEEEpeerreviewmaketitle

\section{Introduction}

Robotic manipulation in unstructured environments requires the sensing ability to perceive, interpret, and respond to rich physical interactions between the robot and its surroundings. Especially, contact-rich manipulation tasks demand on multimodal sensory feedback, including visual, tactile, force, proximity, and temperature cues. Integrating such sensing modalities into robotic end-effectors is a key step toward achieving human-level dexterity and enabling adaptive interaction in real-world scenarios.

Among these sensing modalities, the visual modality is the most widely used, as it effectively captures environmental information such as the size, color, shape, and texture of objects. However, it becomes less effective during contact-rich manipulation due to occlusion. In contrast, tactile sensing enables direct physical interaction, providing reliable information even under visually degraded conditions. Tactile sensing features can be categorized into static (e.g., surface texture, contact shape, indentation depth) and dynamic (e.g., force, shear, torque) components, which can be defined as sub-modality information. They complement vision by providing grounded physical information essential for contact-rich tasks such as object manipulation, exploration, and in-hand adjustment. Thereby, tactile perception can enable the estimation of physical parameters that are otherwise difficult to infer visually, such as mass distribution, friction coefficients, and dynamic response. advanced computer vision techniques. Beyond visual and tactile modalities, proximity sensing plays a critical intermediary role to further bridge the perception-action gap in manipulation tasks. While visual sensing may fail at close range due to occlusions or limited field of view, and tactile sensing requires direct contact which may involve risks, proximity perception offers a buffer zone that enhances precision and safety during fine manipulation. It enables robots to detect approaching surfaces, adjust trajectories, and prepare for contact, improving task reliability in dynamic environments.



Among the diverse array of tactile sensor technologies, vision-based tactile sensors (VBTSs) have emerged as a compelling solution due to their ability to precisely capture contact information by leveraging mature vision system. However, the integration of VBTSs with visual, tactile, and proximity modalities into robotic grippers remains limited by several key challenges, including structural redundancy, sensor miniaturization, mechanical compliance, and the need for real-time multimodal perception software. These limitations hinder the development of intelligent end-effectors capable of operating reliably in contact-rich environments. To address these challenges, we propose \textbf{MagicGripper} in this work, a compact multimodal sensor-integrated gripper that builds upon our previously developed MagicTac technology\cite{fan2024mag}. 
At the core of MagicGripper is the \textbf{mini-MagicTac}, a miniaturized version of MagicTac designed specifically for integration into robotic grippers. Fabricated using multi-material additive manufacturing, mini-MagicTac features a multi-layered grid elastomer architecture. This structural innovation enables the simultaneous perception of multiple sensing modalities, including visual, proximity, and tactile sensing, without requiring additional hardware components.
In summary, MagicGripper addresses the limitations of existing VBTS-integrated grippers through the following contributions:

\begin{itemize}
    \item It introduces mini-MagicTac, which features a multi-layered grid structure embedded in a 3D-printed elastomer. It enhances tactile sensing of both static and dynamic features, while leveraging the design flexibility, manufacturing efficiency and product quality.
    
    \item It integrates visual, proximity, and tactile modalities within a compact gripper by exploiting the reflective and refractive properties of the mini-MagicTac’s grid cells. This enables multi-modal sensing within a single structural unit, eliminating the need for bulky hardware stacking or redundant sensors.
    
    \item It proposes a sensing framework for the MagicGripper that decouples proximity and contact events using channel entropy correlation and grid similarity metrics. This supports seamless transitions between pre-contact, contact, and post-contact phases during manipulation.
\end{itemize}

Through these contributions, MagicGripper demonstrates multimodal perception can be embedded within robotic end-effectors to enhance interaction autonomy, which also offers a scalable path toward developing next-generation sensor-rich robotic manipulators for contact-aware intelligence.

\section{Related Work}
\label{related}

\subsection{Sub-modal Tactile Sensing in VBTSs}

Sub-modal tactile sensing refers to the ability of a sensor to capture distinct components within the tactile modality, such as static texture, or dynamic force. For VBTSs, this capability is closely tied to their sensing mechanisms \cite{fan2024crystaltac}, which define how physical tactile interactions are optically encoded into image data. One common approach is the \textit{intensity mapping method} (IMM), typically used in GelSights \cite{yuan2017gelsight, gomes2020geltip, lambeta2020digit, padmanabha2020omnitact}. These sensors utilize reflective surface layers to capture static contact information, including fine surface geometry and local contact depth.
In contrast, the \textit{marker displacement method} (MDM) captures dynamic contact information by tracking the movement of embedded marker patterns \cite{li2023marker}. Most MDM-based VBTSs use single-layer marker patterns, which are either embedded within the elastomer \cite{yang2021enhanced} or located on its surface \cite{lepora2022digitac}. Also, some works explored multi-layer marker designs—such as GelForce \cite{sato2009finger}, applying dual layers to estimate traction fields, and ChromaTouch \cite{lin2019sensing, lin2020curvature}, which employs subtractive color mixing for enhanced tactile feature detection. Hybrid designs combining IMM and MDM have also been proposed \cite{taylor2022gelslim, do2023densetact, zhang2023gelstereo}, enabling simultaneous perception of static and dynamic tactile features. However, these designs often inherit the limitations of both methods—for instance, reflective layers are prone to abrasion, while marker patterns may occlude underlying features, reducing mapping fidelity.

\subsection{Multi-modality Sensing of VBTSs}

Beyond improving sub-modal sensing within the tactile modality alone, researchers have explored \textit{modality fusion methods} (MFMs) \cite{fan2024crystaltac} to leverage the complementary strengths of multiple sensing modalities, aiming to enhance perception capabilities by integrating visual, tactile, proximity, or even thermal features within a unified sensing architecture.

For example, FingerVision \cite{yamaguchi2016combining, yamaguchi2021fingervision} and ViTacTip \cite{fan2024vitactip} incorporate visual features into marker-based tactile sensing using a transparent elastomer. Notably, ViTacTip employs a generative adversarial network (GAN) to switch between fused and single-modality data representations.
UVtac \cite{kim2022uvtac} and SpecTac \cite{wang2022spectac} introduce ultraviolet (UV) illumination to activate marker-based tactile sensing. These designs leverage reflective coatings and transparent elastomeric skins to extract fine textures and additional visual features.
STS \cite{hogan2021seeing} and VisTac \cite{athar2023vistac} utilize internal lighting and mirror coatings to enable modality switching between purely visual and tactile modes, while TIRgel \cite{zhang2023tirgel} relies on camera focus adjustment to alternate sensing modes.
SATac \cite{10610373} integrates a thermoluminescent layer with marker-based sensing, enabling simultaneous detection of tactile and thermal signals. Similarly, M3Tac \cite{10682561} combines near- and mid-infrared cameras to perform multispectral imaging, allowing multimodal perception of texture, force, proximity, and temperature.


\section{Methodology}
\label{Design-Fabrication}

In this section, we introduce the multi-layered grid structure embedded in the \textbf{mini-MagicTac}, the core sensing component of \textbf{MagicGripper}. We first explain its working principles for enabling multi-modal sensing, including visual, proximity, and tactile modalities. Then, we describe the overall structural design of the complete MagicGripper system.

\begin{figure}[!htbp]
    \centering
    \includegraphics[width = 0.95\hsize]{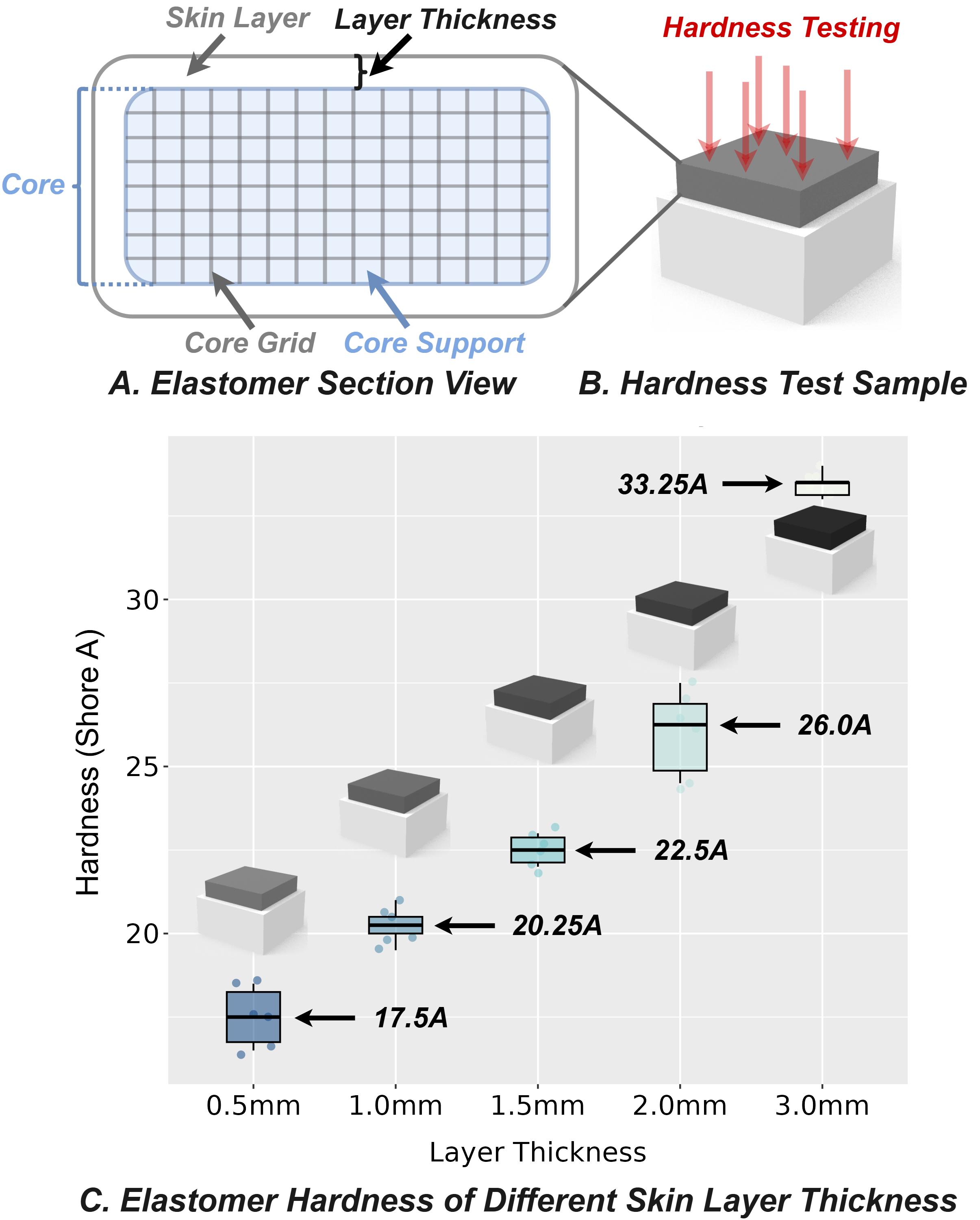}
    \caption{A: Printed elastomer consists of the external skin layer and the internal core embedded with multi-layer grid. B: Samples with various structural configurations are printed for hardness evaluation. C: Hardness can decrease from around 33A to 17A when skin layer varies from 3mm to 0.5mm thick.}
 \label{elastomer_stiffness}
\end{figure}

\subsection{Multi-layer Grid embedded in mini-MagicTac}

As proposed in our previous work \cite{fan2024mag}, the multi-layer grid is fabricated using a multi-material additive manufacturing technique known as PolyJet printing (PP). This integral printing process simultaneously constructs the grid skeleton and the infilled core using different materials, eliminating the need for post-print assembly. It addresses several limitations of conventional manufacturing methods, such as the reliance on manual skills and experience, restricted design flexibility, and inconsistent assembly quality. The grid cell dimensions are fully customizable, typically ranging from 0.6 mm to 1 mm to balance elastic deformability with manufacturing precision.
Specifically, Agilus30 Clear is used for internal grid skeleton due to its rubber-like properties, which offer high deformability, durability, and robustness during physical contact events. A soft support material, SUP706, is employed to fill the interior of the grid cell. This support material enhances the grid's structural integrity and resilience while also reducing its overall hardness.
It is worth noting that Agilus30 Clear is transparent, while SUP706 is translucent. As a result, the multi-layered grid exhibits partial light transmittance, making it potential for use as VBTS elastomer, which has been demonstrated in \cite{fan2024crystaltac}. This optical characteristic also opens up opportunities for enabling multimodal sensing capabilities.


As illustrated in Fig.~\ref{elastomer_stiffness} (A), an external skin will be fabricated around the printed elastomer, where multi-layer grid is embedded inside as the core. This hybrid structure brings customisability, such as elastomer hardness, where VBTSs typically use elastomers with a hardness ranging from 5-20A\cite{yuan2017gelsight}. To demonstrate such capability, several samples were printed for hardness testing (Fig.~\ref{elastomer_stiffness} (B)). When the height of the elastomer is kept constant (10mm), the thickness of the skin layer and the height of the core are inversely proportional, where the thicker the skin or the thinner the core, the closer the elastomer hardness is to the properties of pure Agilus 30 (30A), which is proved by test results in Fig.~\ref{elastomer_stiffness} (C). It is also important to note that when the thickness of the skin layer is constant, the height of the core inside will also have an effect on the elastomer hardness, where the thicker the core, the smaller its hardness. As summarized in Table.~\ref{hardness table}, hardness of multi-layer grid can be customized by adjusting the dimension of each sub-component in printed elastomer. 

However, it should be noted that the above adaptations still have to follow certain design constraints. Although a smaller thickness of the skin layer can lead to softer properties and better tactile sensitivity, it also makes it more vulnerable to abrasion during physical contact with external objects. In comparison, a thicker skin enhances the sensor's durability, but reduces the height of the multi-layer grid core when the total height of the elastomer remains constant, thus limiting sub-modal tactile sensing capability. After comprehensive analyses, a skin thickness beetween 0.5-1mm and a core height of about 5mm are suitable for embedded grid in mini-MagicTac.


\begin{table}[!htbp]
\renewcommand{\arraystretch}{1.5}
\caption{Printed Elastomers with Adjustable Structures \& Hardness}
\begin{tabular}{cccc}
\hline
\textbf{Elastomer Height} & \textbf{Skin Thickness} & \textbf{Core Height} & \textbf{Test Hardness} \\ \hline
10mm                            & 3mm                             & 4mm                           & 33.25A                         \\
10mm                            & 2mm                             & 6mm                           & 26.0A                          \\
10mm                            & 1.5mm                             & 7mm                           & 22.25A                         \\
10mm                            & 1mm                             & 8mm                           & 20.25A                         \\
10mm                            & 0.5mm                             & 9mm                           & 17.5/16.7A                     \\
2mm                             & 0.5mm                             & 1mm                           & 30.5A                          \\
5mm                             & 0.5mm                             & 4mm                           & 18.2A                          \\
15mm                            & 0.5mm                             & 14mm                          & 15.17A                         \\
20mm                            & 0.5mm                             & 19mm                          & 14.2A                          \\ \hline
\end{tabular}
\label{hardness table}
\end{table}

\begin{figure}[]
    \centering
    \includegraphics[width = 0.9\hsize]{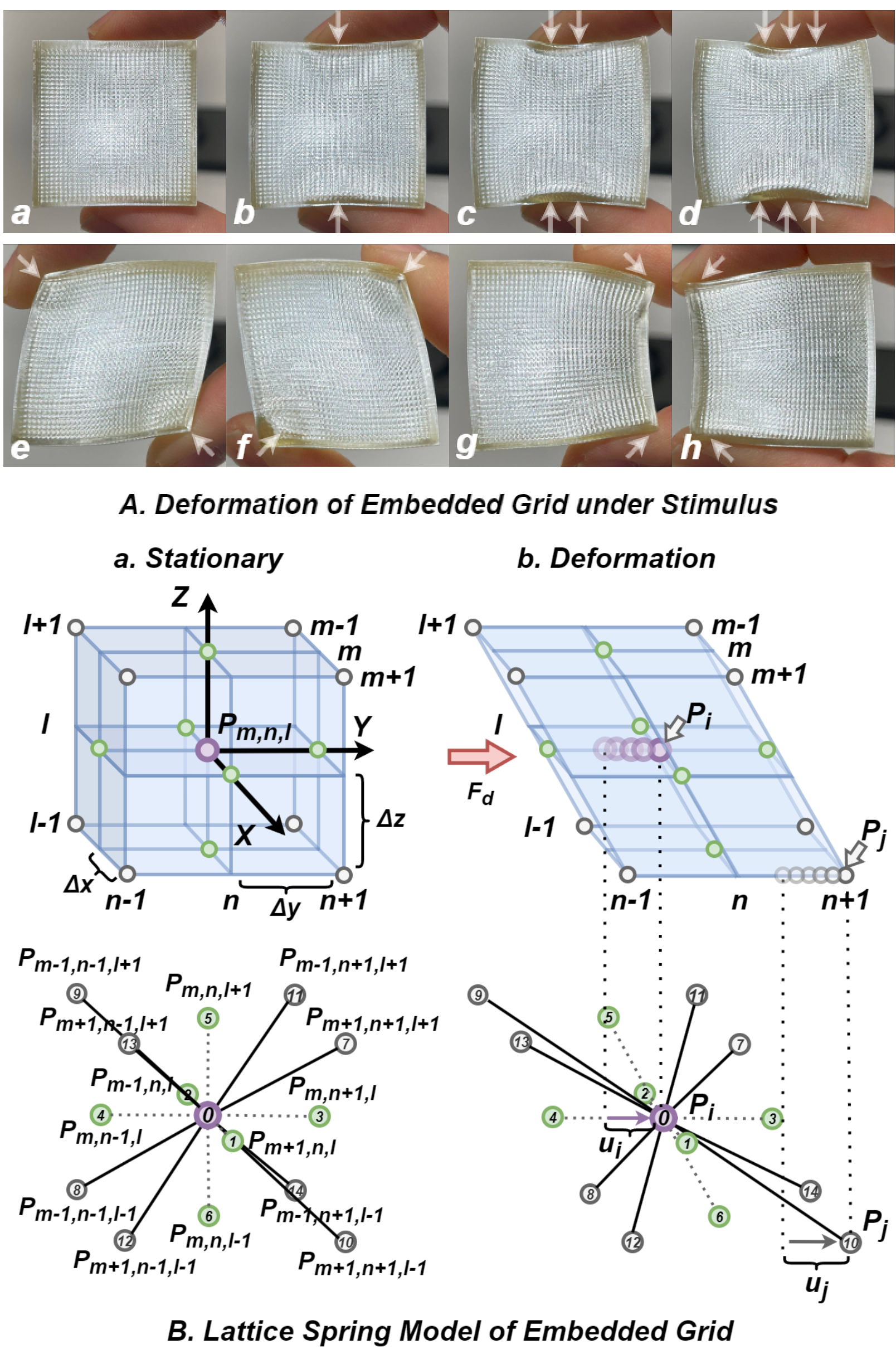}
    \caption{A: Distribution of grid cells (a) will change at the neighbor area around the contact point, which is sensitive to different levels of force (b,c,d), directions (e,f) and rotation (g,h). B: Lattice spring model (LSM) of multi-layer grid in stationary and deformation states.}
    \label{lsm model}
    \vspace{-1em}
\end{figure}

\subsection{Sensing Property of Multi-Layer Grid}

Multi-layer grid structure offers two unique properties: (1) High density and uniform cell distribution enables it with 3D tactile response for both local/global contact and static/dynamic stimulus; (2) The deformability allows it to be sensitive to tactile information while also whose optical property offering visual and proximity sensing capabilities. Below we investigate deformation property of multi-layer grid structure as well as its optical analyze to show its potential in multi-modality sensing:

\subsubsection{Deformation Analysis}
As shown in Fig.~\ref{lsm model} (A), grid structure can map various sub-modal tactile information in cell distribution, such as stimuli in different force, direction and rotation. When there is no external contact, the grid cells remain uniformly distributed throughout the interior of the elastomer (Fig.~\ref{lsm model} (A.a)). When stimuli are applied to the elastomer surface with a light force, the grid layers closest to the contact surface starts to deform elastically, in the direction of the applied force (Fig.~\ref{lsm model} (A.b)). As the applied force is gradually increased, two significant changes occur, the first is an increase in its deformation magnitude (Fig.~\ref{lsm model} (A.c)), and the second is an increase in a deformation region around the contacting surfaces (Fig.~\ref{lsm model} (A.d)). The above scalling law holds true when the applied force changes the direction of application (Fig.~\ref{lsm model} (A.e/f)) and the angle of rotation (Fig.~\ref{lsm model} (A.g/h)), demonstrating its isotropic nature.

In order to illustrate the above properties in more detail, we introduce lattice spring model (LSM) for embedded grid, which is presented in Fig.~\ref{lsm model} (B). The exampled structure consists of a 2×2×2 grid of eight flexible cells with size $(\Delta x, \Delta y, \Delta z)$, which can be arranged into three plane layers: the upper plane layer $(l+1)$, the middle plane layer $(l)$, and the lower plane layer $(l-1)$. These 8 adjacent cells can form a minimal lattice unit of representative elementary volume (REV) for the entire multi-layer grid in printed elastomer, whose central point $P_{m,n,l}$ (purple) is denoted as local coordinate origin for frame defination. In the stationary state (Fig.~\ref{lsm model} (B.a)), 4 structure nodes (green) and 8 diagonal nodes (white) are selected for spring connection, whose corner points on each layer overlap each other. Once contact stimuli $F_d$ occurs and grid deformation is generated, each two linked points in LSM, $P_i$ and $P_j$, will lead to displacements $u_i$ and $u_j$, as illustrated in Fig.~\ref{lsm model} (B.b). Obviously, the different properties of stimuli $F_d$, including its magnitude, direction and rotation, directly affect the scale of $u_i$ and $u_j$ and the positions of $P_i$ and $P_j$ when the steady state is finally reached. Whereas in a real mini-MagicTac, the above REV of LSM would be discretely distributed throughout the entire elastomer interior, resulting in a continuous representation of the sub-modal information in tactile modality.

In summary, the mutual interaction between the multi-layer grid and the external stimulus can be modeled as soft body deformation under different force, which is commonly implemented by finite element method (FEM) in corresponding research. Then, REV of LSM can be regarded as the visualised FEM model of multi-layer grid, whose internal cell distribution can reflect stress and strain in printed elastomer, thus mapping out useful sub-modal tactile information.

\begin{figure}[]
    \centering
    \includegraphics[width = 1\hsize]{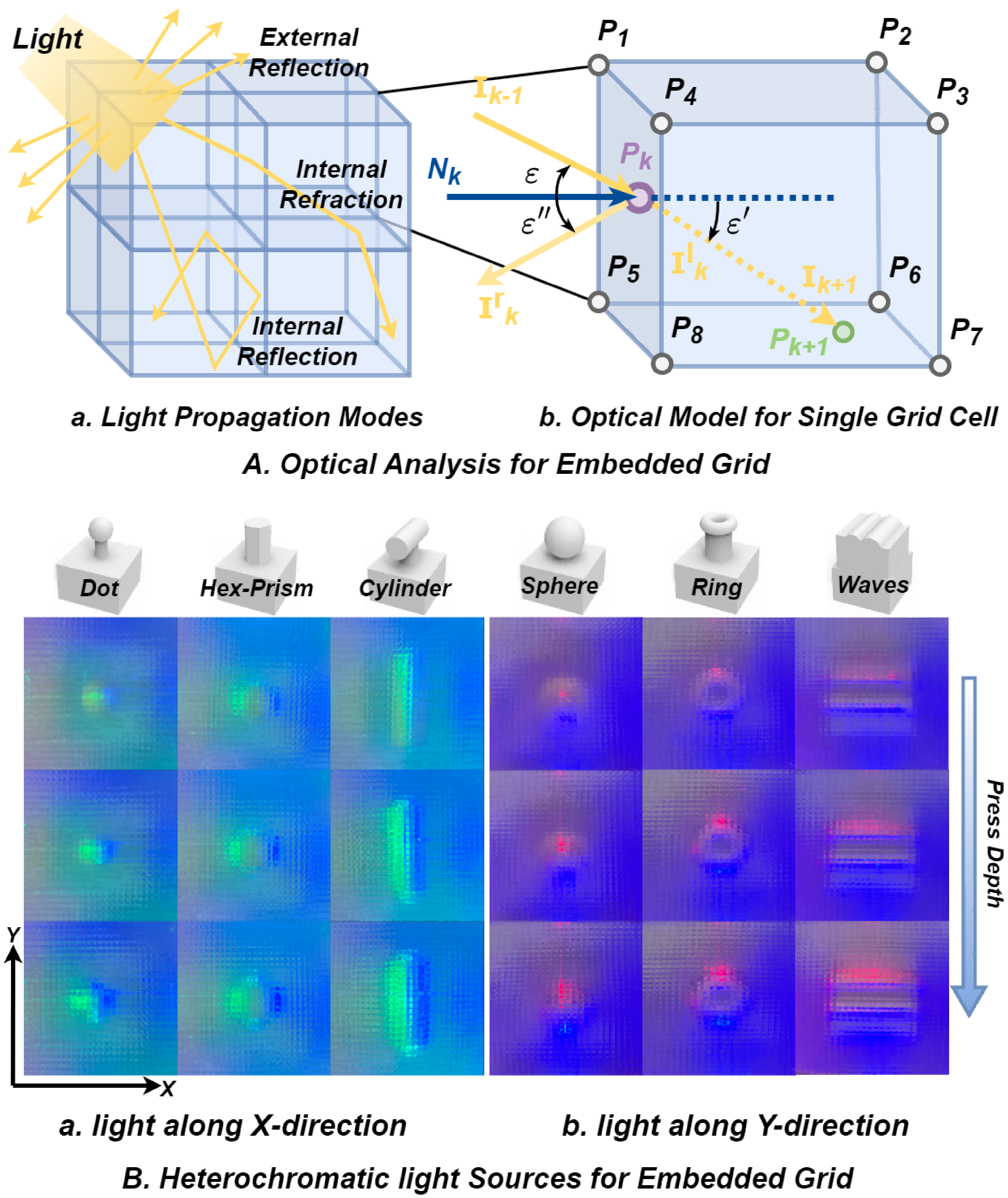}
    \caption{A: Three light propagation modes are coupled inside the grid where internal reflection enhance deformation mapping while internal refraction helps external visual features pass through.  B: From the real test, both visual and tactile features of contact objects can be captured by multi-layer grid during before and after contact happen.}
    \label{optical model}
    \vspace{-1em}
\end{figure}

\begin{figure*}[!htbp]
    \centering
    \includegraphics[width = 1\hsize]{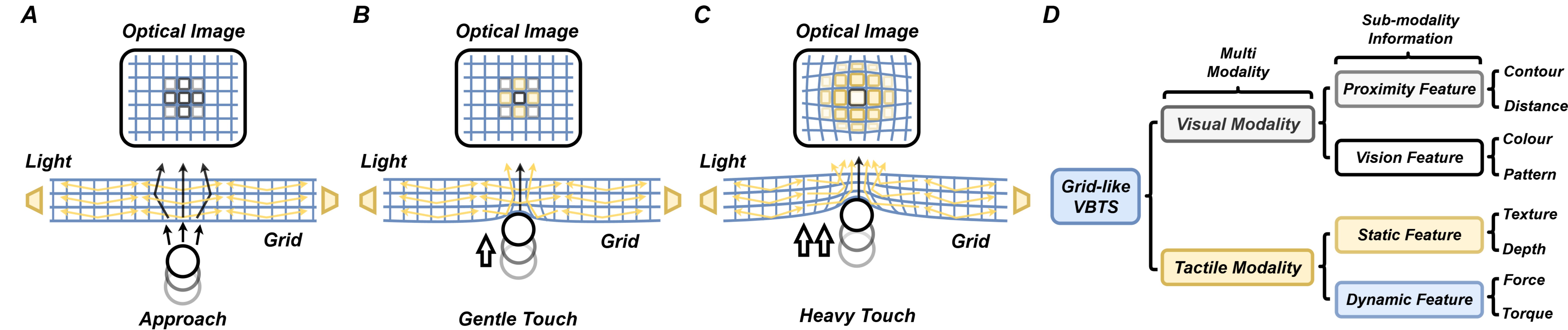}
    \caption{Multi-modality sensing principle of multi-layer grid. A: As the object approaches, the overall grid structure remains unchanged, but the object's proximity and vision features can pass through the grid through internal refraction, and finally be presented in the optical image. B: When the object gently touches the grid, its vision feature will be clearer and only the surface skin will be deformed, thus causing internal reflection of the neighboring grid cells, which is expressed in the optical image as a brightening of the contact area for static tactile feature. C: When heavier contact occurs, the grid structure starts to deform in multiple layers, which helps the optical image to capture the dynamic tactile feature. D: Multi-layer grid can capture both visual and tactile modalities, including sub-modal information of proximity, vision, static and dynamic contact features.}
    \label{multi modality}
\end{figure*}

\subsubsection{Optical Analysis} 
Based on REV of LSM, an optical analysis can be also conducted, as illustrated in Fig.~\ref{optical model} (A). Now suppose a ray of light is incident in it at a random angle, then three forms of optical propagation modes will be generated, namely (1) external reflection, (2) internal reflection, and (3) internal refraction:

\begin{itemize}
    \item Both external reflection and internal reflection occur when the incidence angle $\epsilon$ of light $I_{k-1}$  is too large when passing through the sidewalls of a grid cell ($P_{1},P_{4}, P_{5}, P_{8}$). The external reflection component causes only a portion of light $I_{k+1}$ can enter grid cells while others $I_{k}^{r}$ will be reflected out with same angle $\epsilon''$, especially if the light source is arranged at an angle other than perpendicular to the grid cell sidewalls. Similar to effect of total internal reflection (TIR), internal reflection encourages light $I_{k}^{r}$ propagate inside neighbored grid cell, which will be more pronounced when undergoing elastic deformation. It gathers light in deformed grid cells thus make it appear brighter.
    
    \item Internal refraction helps light $I_{k-1}$ with a small incidence angle $\epsilon$ to successfully pass through the sidewalls of the grid cell with angle $\epsilon'$, and the more perpendicular the incidence attitude $\epsilon$, the more light $I_{k}^{l}$ could be successfully incident. At the same time, it also provides an opportunity for the light rays $I_{k+1}$ that undergo internal reflection to penetrate out of the current grid cell, helping camera to capture such light accumulated inside it. 
    
\end{itemize}

In order to verify the above optical analyses, we performed experiments using real printed elastomers and various contact samples, as shown in Fig.~\ref{optical model} (B). In the initial state, we set the height of the elastomer as the Z-axis, and the camera is arranged in the positive direction of the Z-axis, where the X- and Y-axes are parallel to multi-layer grid in elastomer. Then, the contact sample is gradually approaching it from the negative direction of the Z-axis until they touch. In order to minimise the external reflection, we choose to place the light source at the side of the elastomer. In addition, two groups of heterochromatic light sources, blue and green, violet and red, were arranged along the X and Y directions to visualise the effect of internal refraction and internal reflection better. 

The experiment results are summarized in Fig.~\ref{optical model} (B.a/b). When contact is just starting to occur, only the top of the sample will be in contact, which leads to a small number of grid cells will experience internal reflection and appear brighter. As the press depth gradually increases, the multi-layer grid deforms along the contour and curvature of the contact sample surface to a greater degree, thereby more grid cells undergo internal reflection. By comparing the Fig.~\ref{optical model} (B.a/b), the brightness distribution of grid cells caused by sample contact coincides with the arrangement of light source, which proves that the internal refraction does play a role in the XY direction. Another phenomenon confirms it when the contact just occurred, the outline of the entire sample below can also be faintly seen, which indicates that internal refraction also exists in the Z direction.

To summarize, the multi-layer grid can be regarded as a transparent elastomer embedded with translucent mesh segments. It allows light to pass through, while also introducing a discrete phenomena of total internal reflection (TIR) in uniformly distributed grid cells, whose geometry and intensity can map the corresponding interaction pattern.

\subsubsection{Multi-modality Sensing Principle of Multi-layer Grid} 
Based on the results of deformation and optical property analysis, the multimodal sensing principle of the multi-layer grid structure in MagicTac is illustrated in Fig.~\ref{multi modality}.

Consider a printed elastomer embedded with a multi-layer grid, illuminated by side-mounted light sources and observed by a top-mounted camera. When an object approaches from below without making physical contact (Fig.~\ref{multi modality} (A)), the interior of the grid maintains a stable horizontal illumination pattern, resembling the effect of TIR. At the same time, visual and proximity-related features of the object can be captured through vertical internal refraction as light passes through the grid layers and into the camera's field of view. In this scenario, the captured features depend on the object's distance from the sensor. At greater distances, the visual signal is dominated by \textbf{coarse proximity features}, such as blurred contours or an approximate depth estimation. As the object moves closer, \textbf{finer vision details}, such as surface patterns and color distribution, become more prominent. These optical behaviors form the foundation of multi-layer grid's \textbf{visual modality} sensing capabilities.


In contrast, when an object comes into contact with the printed elastomer, it inevitably alters the embedded grid structure. If the contact is light (Fig.~\ref{multi modality} (B)), only the outer skin deforms while the internal grid remains largely unaffected. This localized surface deformation causes a change in internal reflection within the grid cells near the contact site, resulting in brighter regions within optical image. The intensity and spatial distribution of these bright areas correspond to the texture and depth of the contact, thereby capturing \textbf{static tactile features}.
As the contact force increases (Fig.~\ref{multi modality} (C)), deformation propagates into the inner grid layers. At this stage, the applied force and torque are translated into internal stress and strain within printed elastomer. These mechanical changes are reflected in the obvious patterns of deformed grid, which also mapped as broader variations in the optical image. This enables the extraction of \textbf{dynamic tactile features}, such as force magnitude, contact direction, and interaction dynamics.


Building on the above analysis, the multimodal sensing principle of the multi-layer grid can be clearly summarized by the integrated framework shown in Fig.~\ref{multi modality} (D). This structure enables the fusion of visual and tactile modalities, each encompassing multiple sub-modalities. The visual modality operates both before and during contact, capturing proximity and vision features such as object contours, distance, color, and surface patterns. In contrast, the tactile modality is primarily activated upon contact, enabling the perception of static and dynamic tactile features, including texture, contact depth, applied force, and torque within the touch region.


\begin{figure}[]
    \centering
    \includegraphics[width = 1\hsize]{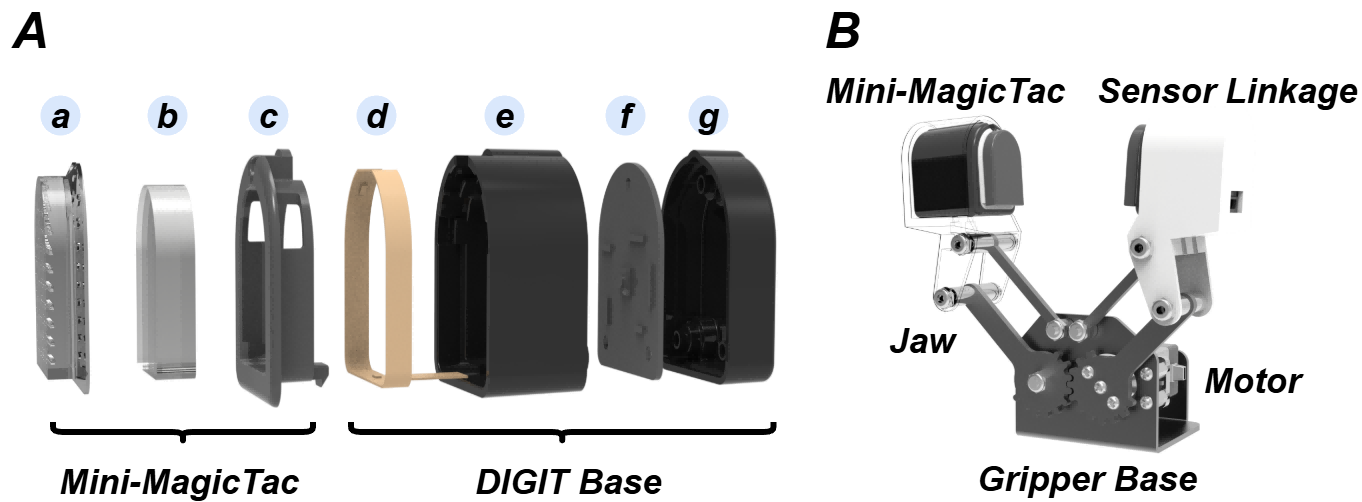}
    \caption{Hardware design of Mini-MagicTac and MagicGripper. A: (a) printed elastomer with multi-layer grid; (b) sensor lens; (c) sensor base; (d) LED strip; (e) DIGIT upper base; (f) camera; (g) DIGIT lower base. B: MagicGripper consists of two motor-driven fingers fixed with Mini-MagicTac.}
    \label{hardware design}
    \vspace{-1em}
\end{figure}

\subsection{Hardware Design of MagicGripper}

To demonstrate such multi-modal sensing capability in real applications, we proposed the hardware design of MagicGripper, as illustrated in Fig.~\ref{hardware design}. Attribute to the design flexibility offered by integral printing technique, multi-layer grid is versatile for a diverse range of VBTS structures and robotic systems. Here, we firstly introduce hardware details of Mini-MagicTac, a finger-shaped contact module which can fit DIGIT's base unit\footnote{https://github.com/facebookresearch/digit-design} properly, as shown in Fig.~\ref{hardware design} (A). Its compactness enhances the adaptability to robotic manipulation, then a MagicGripper can be proposed with two motor-driven finger architecture, where two Mini-MagicTac can be installed (Fig.~\ref{hardware design} (B)). 
The gripper opens and closes via a motor mounted at the base, which drives both fingers to rotate symmetrically in opposite directions. This motion changes the distance between the two mini-MagicTac units while keeping them parallel, allowing for stable, contact-rich manipulation.

\begin{figure}[!htbp]
	\centering
	\includegraphics[width = 0.95\hsize]{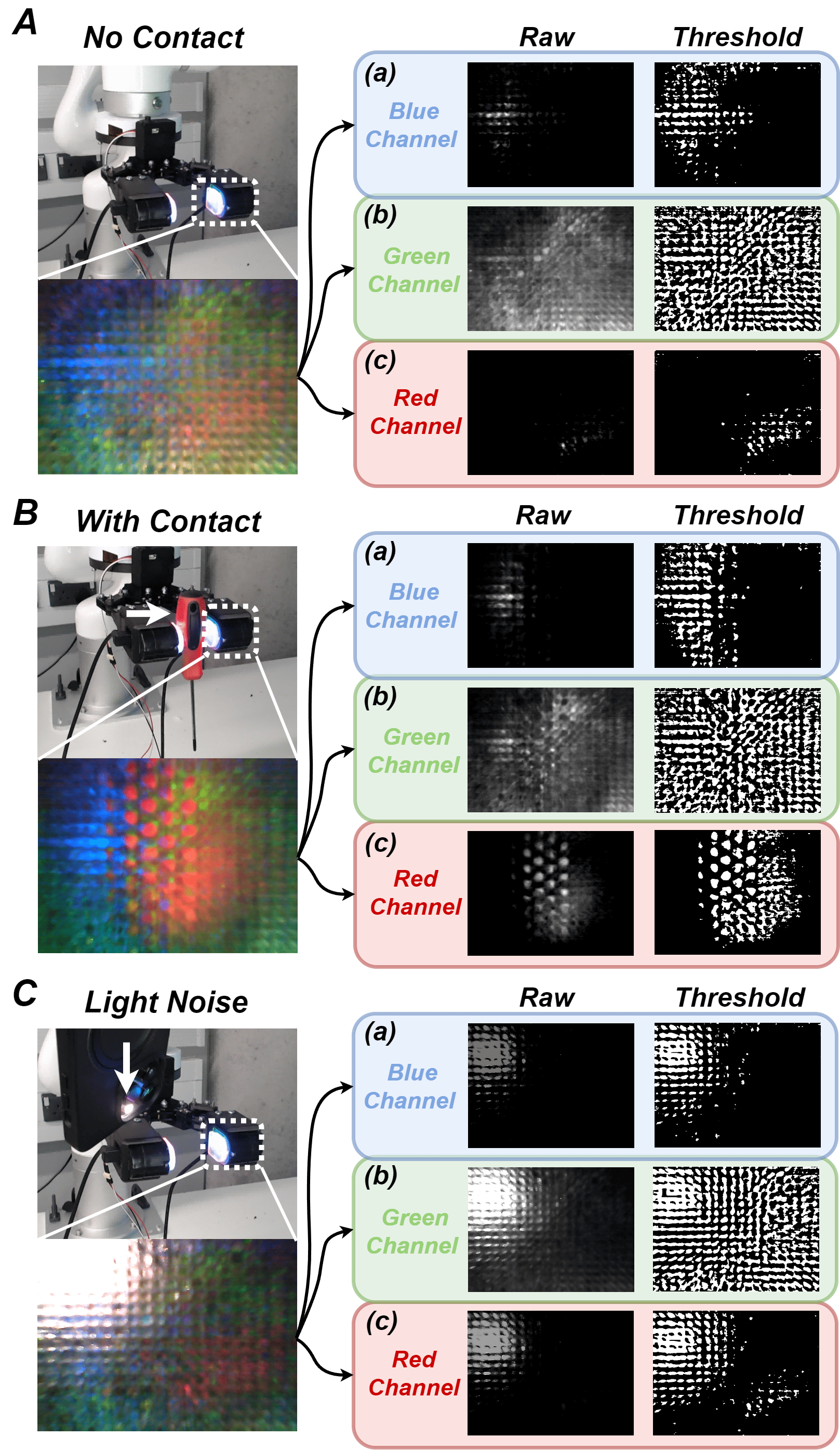}
	\caption{Sensing property analysis of Mini-MagicTac. With heterochromatic light sources of different angles, each channel of raw image in Mini-MagicTac captures distinct features, which depends on whether there is proximity/contact (A, B) or external light interference (C).}
	\label{sensing property}
\end{figure}

\begin{figure}[!htbp]
	\centering
	\includegraphics[width = 1\hsize]{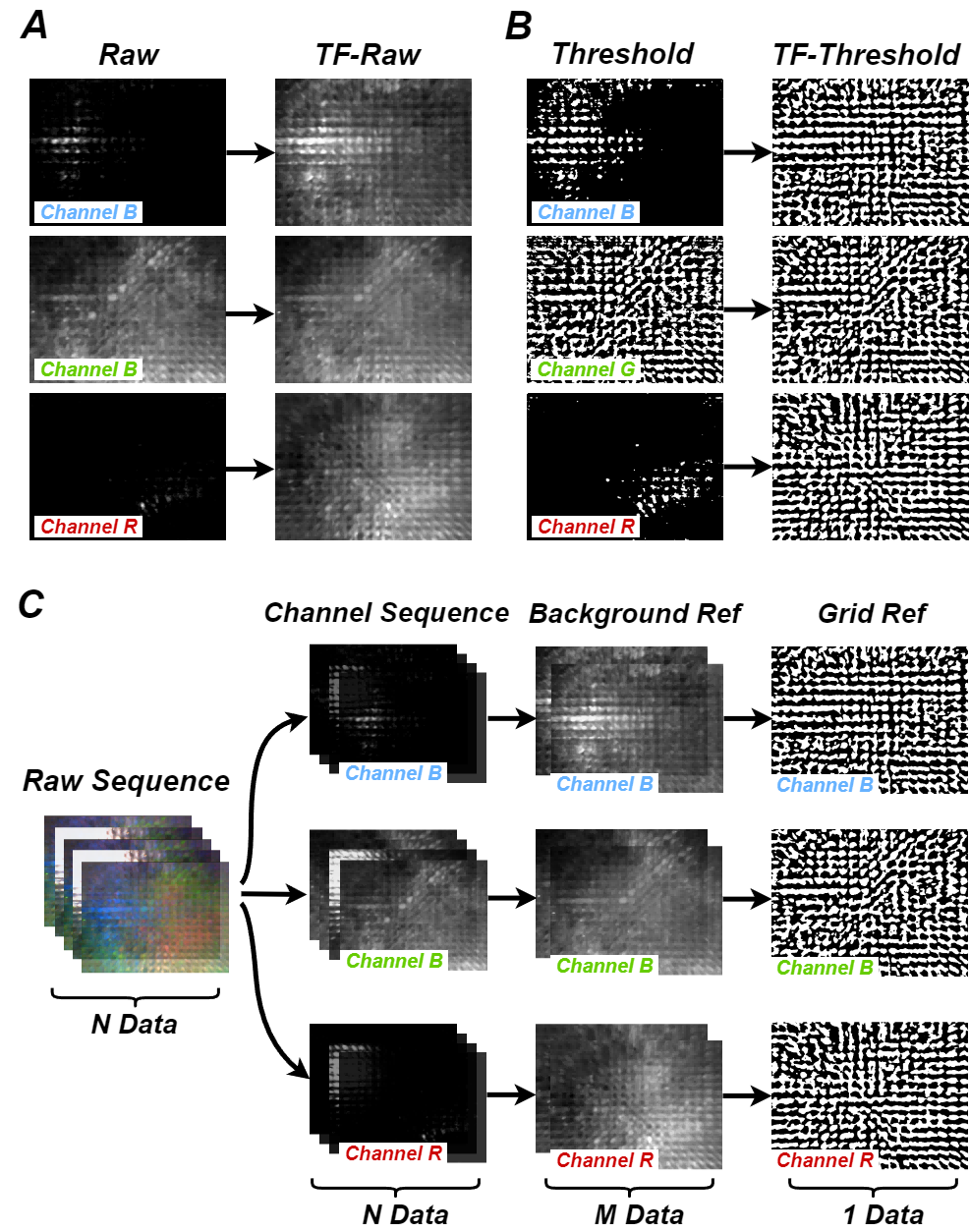}
	\caption{A/B: Temporal fusion (TF) helps for both background fusion and global grid recognition. C: For reference acquisition, temporal fusion is applied to the raw image by channel according to the ratio of N:M, resulting in M background references and one grid reference.}
	\label{temporal fusion}
\end{figure}

\subsection{Proximity and Contact Detection Algorithm}

\begin{figure*}[]
	\centering
	\includegraphics[width = 1\hsize]{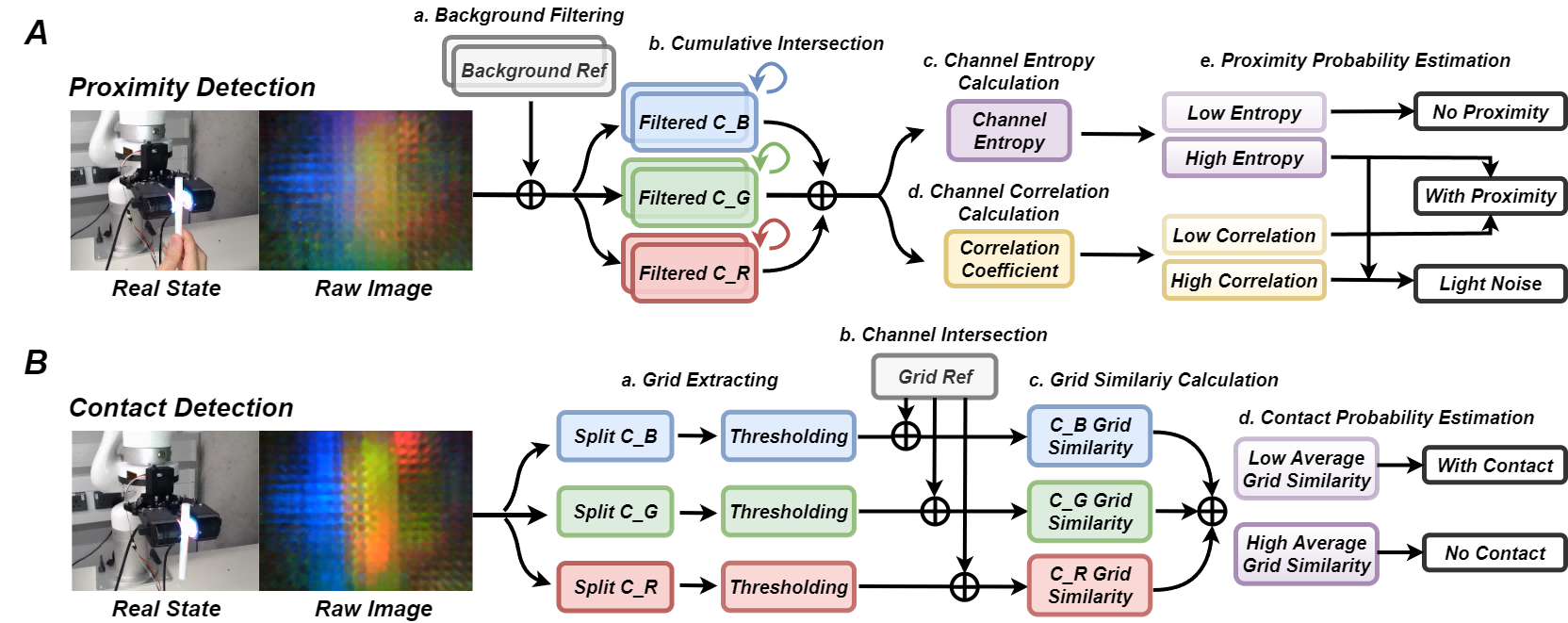}
	\caption{A: Proximity detection framework: (a) first filter the background using the collected reference; (b) apply cumulative intersection for each channel, followed by calculating entropy (c) and correlation coefficient (d) to determine proximity probability. B: Contact detection framework: (a) perform grid extraction for each channel; (b) apply the collected grid reference for the intersection process, whose results are used to calculate grid similarity (c), and then complete the contact probability estimation.}
	\label{proximity contact framework}
\end{figure*}

Based on the multimodal sensing principle of the multi-layer grid, as introduced in Fig.~\ref{multi modality}, we design proximity and contact detection algorithms tailored for the MagicGripper. These capabilities are critical for enabling responsive and adaptive robotic manipulation.

\subsubsection{Sensing Property Analysis of Mini-MagicTac}
To firstly capture sensing characteristics of the mini-MagicTac using DIGIT’s base unit, we analyzed its raw image data which involves with red, green, and blue lighting. Channel-wise image decomposition was performed to assess optical response and sensing variation, as illustrated in Fig.~\ref{sensing property}.


Our first observation is that channel behavior varies significantly: the blue and red channels show uneven spatial illumination, while the green channel provides more consistent coverage across the image (Fig.~\ref{sensing property} (A)). These differences are influenced by both the contact state and external lighting conditions. As shown in Fig.\ref{sensing property} (B), during contact, the disparity between channels becomes more pronounced. In contrast, under ambient light interference (Fig.\ref{sensing property} (C)), the similarity between channels increases due to noise convergence.
A further key observation arises from thresholding the images: the underlying geometry of the multi-layer grid becomes distinctly visible, particularly in the green channel. While this highlights the optical sensitivity of the grid, it also reveals challenges such as cross-channel variance and robustness to random light noise. These factors must be carefully addressed to ensure reliable data processing in MagicGripper system.


\subsubsection{Temporal Fusion Method for Data Processing}



To extract stable and informative features from each color channel, we introduce a \textbf{temporal fusion (TF)} method that averages raw image data over a defined period to generate representative reference frames, typically during the initialization phase of mini-MagicTac.
As illustrated in Fig.~\ref{multi modality} (A), although illumination may propagate in a variable manner due to internal reflection and refraction, the geometry of the embedded grid still remains static in the absence of contact. This property suggests that the image data should contain consistent global patterns across time, particularly in regions associated with the grid structure.
Experimental results support this hypothesis, as shown in Fig.~\ref{temporal fusion} (A/B). The two visual outputs (\textbf{TF-raw} and \textbf{TF-threshold}) clearly demonstrate stable global features, including well-defined grid structures, validating the effectiveness of the temporal fusion strategy. The TF-raw image is produced by averaging raw image frames over time, which preserves the global illumination pattern while suppressing high-frequency noise caused by ambient light fluctuations or minor vibrations. In contrast, the TF-threshold image applies binary thresholding to each frame prior to fusion, emphasizing high-contrast regions. This sharpens the structural features and improves visual segmentation of the grid geometry.

Based on these outputs, we construct two types of reference masks from the temporal sequence of raw mini-MagicTac images: a \textbf{background reference mask} and a \textbf{grid reference mask}. The background reference mask is used to filter out environmental noise, while the grid reference mask provides a baseline for detecting structural deformation during contact events. As illustrated in Fig.~\ref{temporal fusion} (C) and Algorithm \ref{alg:TF function}, a set of $N$ raw images is first split by RGB channels. Each channel is then processed using a temporal fusion ratio of $N:M$ to generate $M$ background reference masks (where $M \leq N$). The values of $N$ and $M$ are user-defined and can be tuned based on the variability of the lighting conditions, improving robustness to temporal disturbances. In contrast, the grid reference mask is constructed as a single fused image from the complete sequence of $N$ frames. This ensures maximal retention of the static grid geometry and serves as a reliable reference for downstream tasks such as contact detection.

\begin{algorithm}[htbp]
\caption{Temporal Fusion for Data Processing}
\label{alg:TF function}
\KwIn{\\
  $I = \{I_1, ..., I_N\}:$ sequence of raw images \\
  $N=30:$ number of raw images \\
  $M = 3:$ number of background reference masks \\
  $\tau_B=35:$ binary threshold for grid reference mask \\
}
\KwOut{\\
  $B_{ref} = \{B_1, ..., B_M\}$: background reference masks \\
  $G_{ref}:$ grid reference mask
}
\BlankLine
\textbf{Step 1: Channel Decomposition} \\
\For {$I_i \in I$}{
    Split $I_i$ into channels $c \in \{r, g, b\}$: $I_i^r$, $I_i^g$, $I_i^b$
}
\BlankLine
\textbf{Step 2: Background Mask Construction} \\
\For {$j = 1$ to $M$}{
    \For{channel $c \in \{r, g, b\}$}{
        $B_j^c = \text{mean}(\{I^c_{(j-1)\cdot (N/M)+1}, ..., I^c_{j\cdot (N/M)}\})$
    }
    $B_j = \text{merge}(\{B_j^r, B_j^g, B_j^b\})$\\
}

$B_{ref} = \{B_1, ..., B_{j=M}\}$\\
\BlankLine
\textbf{Step 3: Grid Mask Construction} \\
\For {channel $c \in \{r, g, b\}$}{

    \For{$i = 1$ to $N$}{
        Apply Gaussian blur to $I^c_{i}$ to filter noise \\
        $G_i^c = \text{binary threshold}(I^c_{i}, \tau_B)$
    }
    $G^c = \text{mean}(\{G^c_{1}, ..., G^c_{i=N}\})$
}
$G_{ref} =\text{merge}(\{G^r, G^g, G^b\})$\\

\Return {$B_{\text{ref}}$, $G_{\text{ref}}$}
\end{algorithm}

\begin{algorithm}[htbp]
\caption{Proximity Detection Using Channel Entropy and Inter-Channel Correlation}
\label{alg:proximity_detection}

\KwIn{\\
  $F_{\text{curr}}:$ current image frame \\
  $\tau_E = 0.5:$ channel entropy threshold \\
  $\tau_C = 0.2:$ channel correlation threshold \\
}

\KwOut{\\
  Proximity state $\in \{\text{Normal}, \text{Approaching}, \text{Noise}\}$
}

\BlankLine
\textbf{Step 1: Background Mask Generation} \\
Same as Step 2 in Algorithm 1:\\
$B_{ref} = \{B_1, ..., B_M\}$: background reference masks \\
$M = 3:$ number of background reference masks \\

\BlankLine
\textbf{Step 2: Preprocess Current Frame} \\
Split $F_{\text{curr}}$ into channels $c \in \{r, g, b\}$: $F^r$, $F^g$, $F^b$\\
\For {channel $c \in \{r, g, b\}$}{
\For {$j = 1$ to $M$}{
        $\Delta F_j^c = \text{threshold}(F^c - B_{j}^c, 0)$\\


        $F_j^c = F_{j-1}^c  \cap \Delta F_j^c$\\
    }
    $F_{filtered}^c = F_{M}^c$\\
}
$F_{intersection} = \text{merge}(\{F_{filtered}^r, F_{filtered}^g, F_{filtered}^b\})$\\

\BlankLine
\textbf{Step 3: Compute Entropy and Correlation} \\

\For {channel $\{r, g, b\}$}{
$C_{rg} = \text{correlation}(F_{filtered}^r, F_{filtered}^g)$ \\
$C_{rb} = \text{correlation}(F_{filtered}^r, F_{filtered}^b)$ \\
$C_{gb} = \text{correlation}(F_{filtered}^g, F_{filtered}^b)$ \\
}

$E_\text{total} = \text{entropy}(gray(F_{intersection}))$ \\
$C_\text{total} = \text{mean}(C_{rg}, C_{rb}, C_{gb})$ \\

\BlankLine
\textbf{Step 4: State Decision Logic} \\
\If{$E_{\text{total}} < \tau_E$}{
    \Return{Normal}
}
\ElseIf{$E_{\text{total}} \geq \tau_E$ \textbf{and} $C_\text{total} < \tau_C$}{
    \Return{Approaching}
}
\ElseIf{$E_{\text{total}} \geq \tau_E$ \textbf{and} $C_\text{total} \geq \tau_C$}{
    \Return{Noise}
}
\end{algorithm}

\begin{algorithm}[htbp]
\caption{Contact Detection Using Grid Similarity}
\label{alg:contact_detection}

\KwIn{\\
$F_{\text{curr}}:$ current image frame \\
  $G_{\text{ref}}:$ pre-computed grid reference mask \\
  $\tau_B=35:$ binary threshold for grid reference mask \\
  $\tau_G = 0.6:$ grid similarity threshold \\}

\KwOut{\\
 Contact state $\in \{\text{Touched}, \text{Untouched}\}$
}


\BlankLine
\textbf{Step 1: Preprocess Current Frame} \\
Split $F_{\text{curr}}$ into channels $c \in \{r, g, b\}$: $F^r$, $F^g$, $F^b$\\
\For {channel $\{r, g, b\}$}{
Apply Gaussian blur to $F^c$ to filter noise \\
$G^c = \text{binary threshold}(F^c, \tau_B)$\\
}

\BlankLine
\textbf{Step 2: Grid Mask Intersection} \\
Split $G_{\text{ref}}$ into channels $c \in \{r, g, b\}$: $G_{ref}^r$, $G_{ref}^g$, $G_{ref}^b$\\
\For {channel $\{r, g, b\}$}{
$G_{intersection}^c = G_{ref}^c  \cap G^c$\\
}

\BlankLine
\textbf{Step 3: Compute Grid Similarity} \\
\For {channel $\{r, g, b\}$}{
$S_r = similarity(G_{intersection}^r, G_{ref}^r)$\\
$S_g = similarity(G_{intersection}^g, G_{ref}^g)$\\
$S_b = similarity(G_{intersection}^b, G_{ref}^b)$\\

}
$S_\text{total} = \text{mean}(S_{r}, S_{g}, S_{b})$ \\

\BlankLine
\textbf{Step 4: State Decision Logic} \\
\If{$S_{\text{total}} < \tau_G$}{
    \Return{Touched}
}
\ElseIf{$S_{\text{total}} \geq \tau_G$}{
    \Return{Untouched}
}

\end{algorithm}

\subsubsection{Algorithm Framework Design}

Next, we introduce the algorithmic framework for proximity and contact detection, which addresses two challenges: (1) First, distinguishing between external interference and valid proximity information from Mini-MagicTac's raw output data; and (2) Second, extracting contact information without complexity to directly track the multi-layer grid in Mini-MagicTac. 

For proximity detection (Fig. \ref{proximity contact framework} (A)), the raw data undergoes background denoising and then cumulative intersection (Fig. \ref{proximity contact framework} (A.a/b)), aiming to identify common variations across channels through self-iteration. The cumulative number of iterations is M, corresponding to the number of background reference masks. After, the channel information is fed into two computational modules - one for calculating the entropy across channels (Fig. \ref{proximity contact framework} (A.c)) and the other for calculating the correlation coefficients between different channels (Fig. \ref{proximity contact framework} (A.d)). Entropy represents the degree of disorder in an image, which changes through visual modality, when an external object comes close to Mini-MagicTac. However, it can be highly sensitive to light noise, which is why correlation coefficients are introduced as a complementary measurement. As shown in Fig. \ref{sensing property} (C), channel differences will decrease during light noise occurs, which can help to distinguish between noise and the proximity/contact of external objects. For instance, when the entropy starts to change, a low correlation coefficient can indicate that an external object is likely approaching since internal reflection/refraction will guide external visual features and heterochromatic light source towards different directions. Whereas a high correlation coefficient suggests that the entropy change may due to external noise rather than real contact, whose internal refraction introduced in the vertical direction will play a dominant role in all channels of optical image. In addition, if the entropy remains consistently low and stable, the system should determine that the state should be normal (Fig. \ref{proximity contact framework} (A.e)). The details of proximity detection framework is explained in Algorithm \ref{alg:proximity_detection}.

\begin{figure}[!htbp]
	\centering
	\includegraphics[width = 1\hsize]{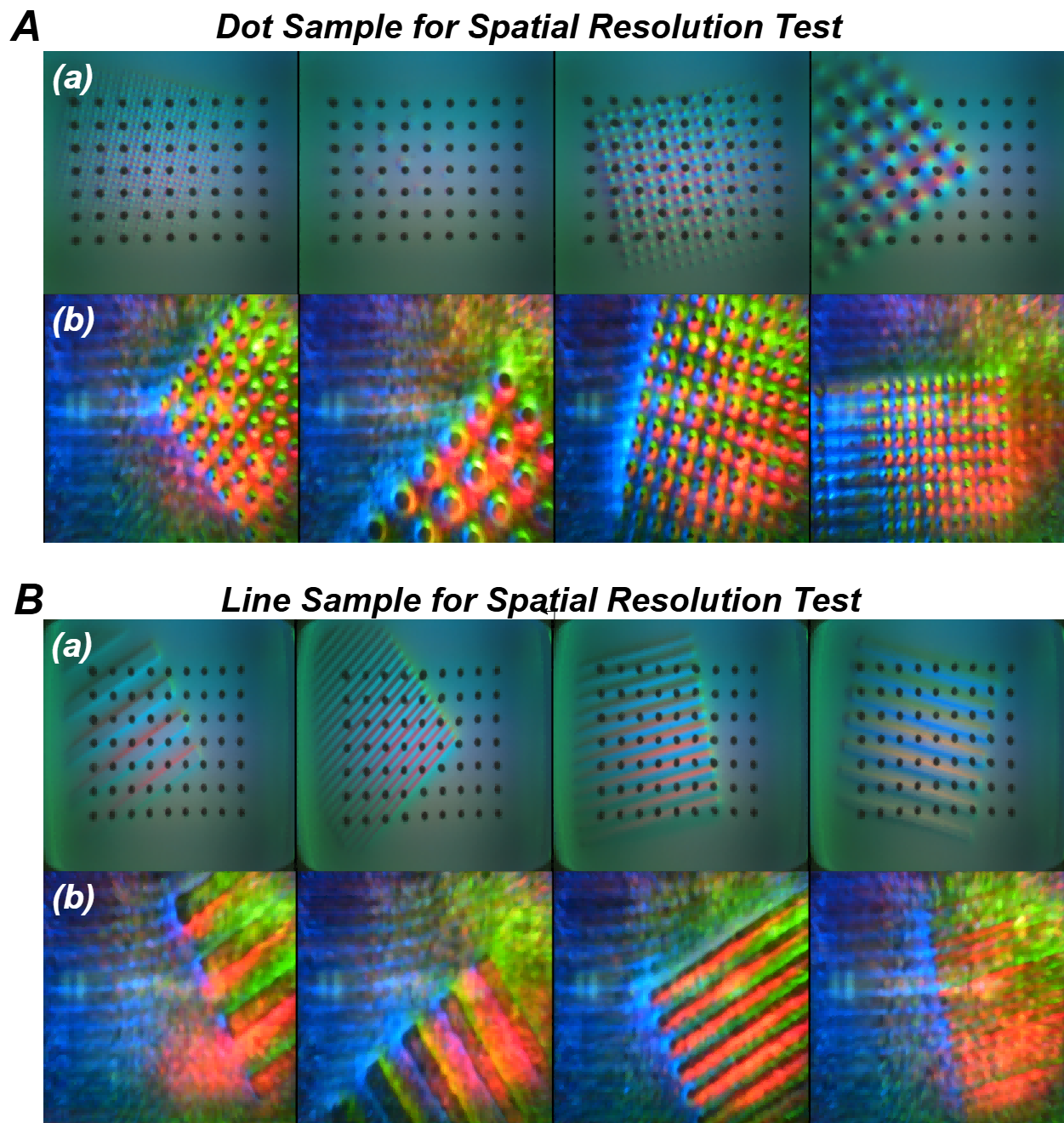}
	\caption{Test samples for spatial resolution. A: Dot sample for spatial resolution test between GelSight (a) and Mini-MagicTac (b). B: Line sample for spatial resolution test between GelSight (a) and Mini-MagicTac (b). }
	\label{spatial resolution}
\end{figure}

\begin{figure*}[!htbp]
	\centering
	\includegraphics[width = 1\hsize]{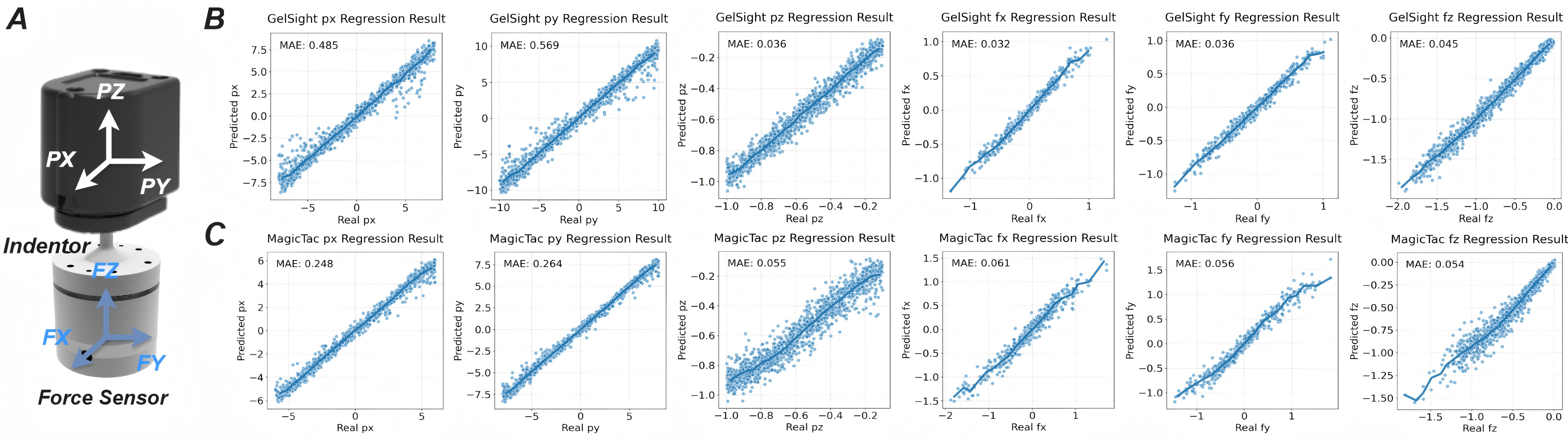}
	\caption{A: Experiment setup for contact localization and force regression, where six pose and force parameters will be recorded (PX, PY, PZ, FX, FY, FZ). B: Contact localization and force regression results of GelSight. C: Contact localization and force regression results of Mini-MagicTac.}
	\label{regression}
\end{figure*}

For contact detection, it highly relies on grid identification, whose pipeline is depicted in Fig. \ref{proximity contact framework} (B). Based on Fig.~\ref{multi modality} (B/C), the external contact will cause the deformation of multi-layer grid, and the heavier the contact, the more pronounced its deformation. In contact detection process, the output data from Mini-MagicTac is firstly split by channel, whose intrinsic geometry of the grid is then extracted through thresholding (Fig. \ref{proximity contact framework} (B.a)). Such geometry is then compared with a previously captured grid reference from Algorithm \ref{alg:TF function} (Fig. \ref{proximity contact framework} (B.b)), whose ratio of the intersecting areas is used to calculate similarity (Fig. \ref{proximity contact framework} (B.c)). A higher similarity between the current channel's grid geometry with related reference indicates less deformation degree, suggesting minimal contact with external objects, otherwise the highly deformed grid will show low-level similarity towards the reference (Fig. \ref{proximity contact framework} (B.d)), resulting in identified contact status. The details of contact detection framework is explained in Algorithm \ref{alg:contact_detection}.

\section{Experimental Evaluation}
\label{Experiment}
In this section, we aim to conduct experimental evaluations for Mini-MagicTac and MagicGripper. 


\subsection{Performance Evaluation of Mini-MagicTac} 

\subsubsection{Spatial Resolution}

Spatial resolution is a representative metric in evaluating the static tactile sensing performance of VBTS. As analyzed in Fig.~\ref{multi modality} (A/B), multi-layer grid can perceive the texture on object surface through both visual clue and internal reflection of grid cell. To demonstrate such capability, a comparative spatial resolution evaluation was conducted between Mini-MagicTac and GelSight, where the latter is a popular commercial product and widely used in fine-texture tactile sensing. As shown in Fig.~\ref{spatial resolution}, two types of test samples are introduced, where dot sample consists of dot arrays with various radius and spacings (convex features), and line sample consists of a row of grilles with various widths and spacings (concave features). Points and lines are the shape primitives that make up the texture of common objects in life, so test results based on them are highly informative. Specifically, each sample has 25 different configurations with feature sizes ranging from 0.2mm to 1.75mm at 0.05mm intervals, enough to cover a wide range of texture sizes. In our test, the sensor was mounted on a robotic arm to perform indentation on corresponding samples, each repeated 100 times. For each trial, the indentation depth was generated within the range of 0.1 mm to 2 mm. The indentation position (x, y) was randomly selected within the sensor’s surface area. Additionally, the sensor was randomly rotated by an angle between -90° and 90° for each trial. This randomized indentation strategy also enhances model robustness by reducing the influence of noise. For both GelSight and Mini-MagicTac, the same amount of data are collected using the identical experimental setup and the models are trained for the identification task using ResNet18 for each of them. 

Here, spatial resolution of sensor is defined as the minimum distance at which two features can be perceived as distinct. The identification accuracy of spatial resolution are summarised in Table.~\ref{spatial resolution table}, which is regarded as a tolerance-window classification problem. For each sensor, a prediction is counted as correct if it falls within $\pm \Delta$ mm of the ground-truth feature size. By varying $\Delta$ from 0.05 mm to 0.50 mm, we assess how accurately each sensor distinguishes feature sizes under different error tolerances. Accuracies are derived from the confusion matrices built on classification results from the indentation experiments. From such results, it can be seen that for dot and line samples, the accuracy of both sensors is 100\% when the resolution is greater than 0.15 mm, until the resolution is as low as 0.1 mm, when the classification accuracy of Mini-MagicTac starts to decrease. In comparison, the GelSight shows a decrease in accuracy when the resolution is 0.05 mm. It can be concluded that the Mini-MagicTac's spatial resolution can reach at least 0.15mm, which is close to the performance of the current SOTA, GelSight.

\begin{table}[!htbp]
\centering
\caption{Accuracy (\%) of GelSight and Mini-MagicTac at different spatial resolution ($mm$)}
\scriptsize 
\begin{tabular}{lcccccccc}
\toprule
Sensor & 0.50 & 0.40 & 0.30 & 0.25 & 0.20 & 0.15 & 0.10 & 0.05 \\
\midrule
GelSight (dot) & 100 & 100 & 100 & 100 & 100 & 100 & 100 & 99.03 \\
GelSight (line) & 100 & 100 & 100 & 100 & 100 & 100 & 100 & 99.72 \\
MagicTac (dot) & 100 & 100 & 100 & 100 & 100 & 100 & 99.67 & 98.33 \\
MagicTac (line) & 100 & 100 & 100 & 100 & 100 & 100 & 99.00 & 98.00 \\
\bottomrule
\end{tabular}\
\label{spatial resolution table}
\end{table}

\subsubsection{Contact Localization and Force Regression}



In dynamic tactile sensing task, the capability to precisely estimate the position and force of contact interaction plays an vital role for VBTS in robotic manipulation. Based on Fig.~\ref{multi modality} (C), multi-layer grid can capture such features in assistance by the deformation of multi-layer grid structure. Therefore, another comparative task is also designed to evaluate the dynamic tactile sensing performance of Mini-MagicTac, whose experiment setup is illustrated in Fig.~\ref{regression} (A). In the experiment, the sensor, mounted on a robotic arm, makes contact with an indenter fixed above a force sensor. The contact force is recorded by the force sensor, while the contact position is derived from the known pose of the robotic arm. Each trial involves a sequence of controlled indentations: the sensor presses down from surface contact by a randomly sampled depth between 0.1 mm and 2 mm, discretized into four levels. At each level, two types of data are collected: (1) normal indentation data, and (2) shear data, obtained by translating the sensor within a random displacement of up to 1 mm in the XY plane while maintaining contact. The sensor is then lifted and re-indented to the next depth, repeating the process four times. This design ensures that some spatial positions are revisited under distinct force conditions—either dominated by normal or shear forces—enriching the diversity of the dataset.

After the data collection and model training (ResNet18), the contact localization and force regression results of GelSight and Mini-MagicTac are summarised in Fig.~\ref{regression} (B/C). In contact localization, Mini-MagicTac's positioning accuracy in the XY plane is about 0.3mm with smaller standard deviation, better than GelSight's 0.5mm. But its estimation accuracy in Z direction is slightly worse, which may potential attribute to the reflective layer provides GelSight advantage in depth perception. For force estimation, GelSight's regression error is between 0.03N and 0.045N, while Mini-MagicTac's is 0.05N-0.06N. A potential reason for this disparity could be that GelSight's black dot marker array can be captured more clearly by the camera imaging than Mini-MagicTac's grid.


\begin{figure}[!htbp]
	\centering
	\includegraphics[width = 1\hsize]{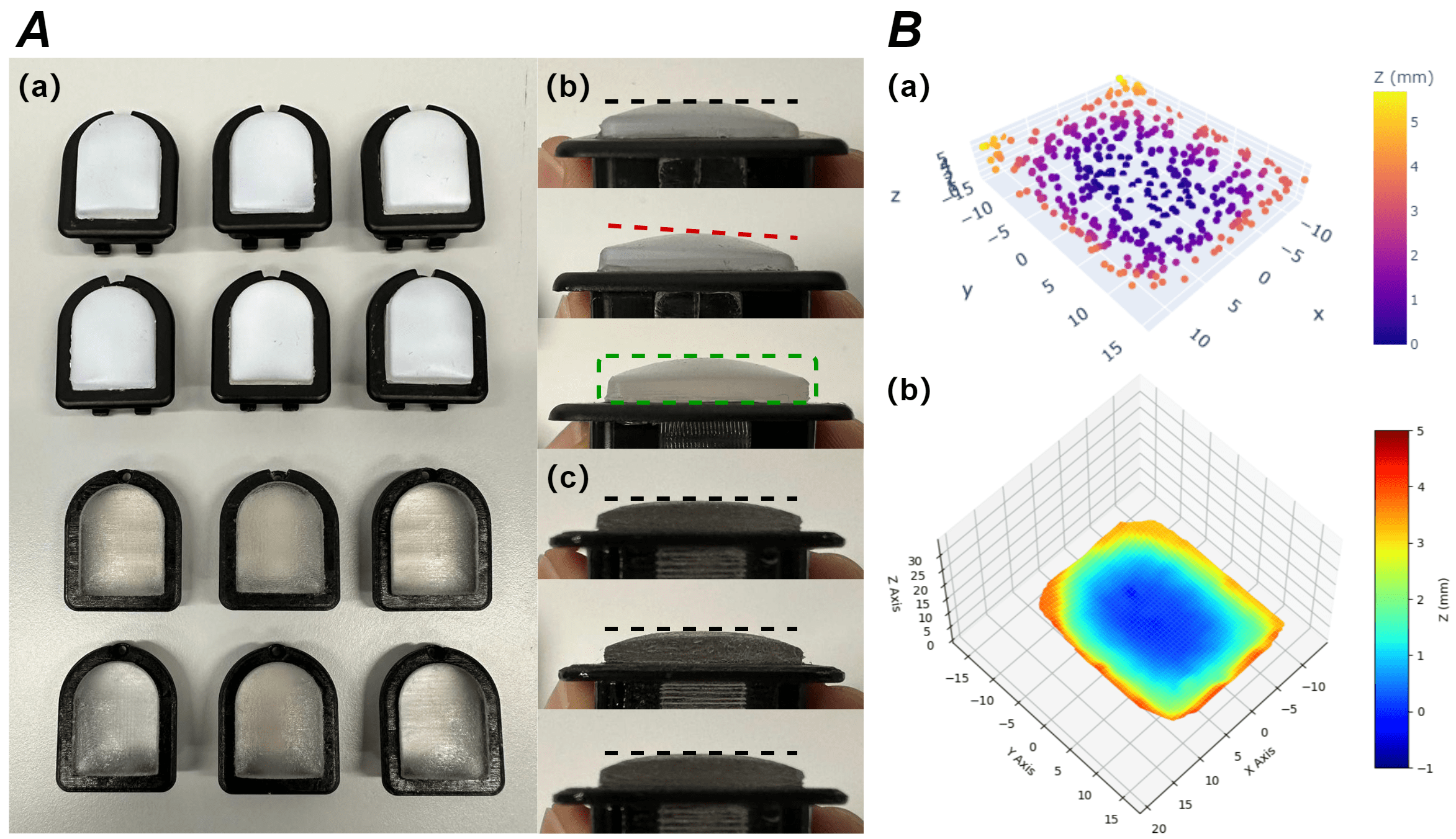}
	\caption{A: (a) Twelves DIGITs and Mini-MagicTacs prepared for manufacturing error evaluation; (b) Significant variation observed in the manufacturing of DIGITs; (c) Mini-MagicTac exhibits more consistent manufacturing quality. B: (a) Multi-point random sampling conducted on each sensor's surface to obtain a point cloud; (b) Final reconstruction results of the sensor surface.}
  \vspace{-0.4cm}
	\label{manufacturing error setup}
\end{figure}

\subsection{Robustness Evaluation of Mini-MagicTac}

To verify the practicality and stability of MagicGripper in complex environments, the robustness of its embedded Mini-MagicTac will be analysed from three perspectives: manufacturing, mechanical, and performance robustness.

\subsubsection{Manufacturing Robustness}

Based on the integral manufacturing approach, Mini-MagicTac in MagicGripper can be fabricated as a single piece, both flexibly and efficiently. To evaluate the quality of integral manufacturing, we conducted a comparative test between DIGIT\cite{lambeta2020digit}, which is produced using traditional manufacturing methods, and shares the same base with Mini-MagicTac. Due to its design specifically for tactile robotic tasks, DIGIT is one of the most popular VBTSs today like GelSight, making it a suitable choice for the control group. As shown in Fig. \ref{manufacturing error setup} (A.a), a total of six DIGITs were used for testing, who were acquired from three separate production batches to mitigate the impact of potential manufacturing deviations in a particular batch. To ensure a fair comparison, we utilized the identical CAD model of DIGIT to fabricate six samples of Mini-MagicTacs. In general, DIGITs have two main problems, one is that the slope of their elastomer surface varies between individuals, as shown by the red line in Fig. \ref{manufacturing error setup} (A.b), which leads to different contact information even under the same contact pose. Another issue is the variation in elastomer thickness between individuals, as shown by the green box in Fig. \ref{manufacturing error setup} (A.b), which causes a mismatch between press depth and contact information. In contrast, Mini-MagicTacs represent better consistency in shape between individuals as shown as black lines in Fig. \ref{manufacturing error setup} (A.c).

\begin{figure}[!htbp]
	\centering
	\includegraphics[width = 1\hsize]{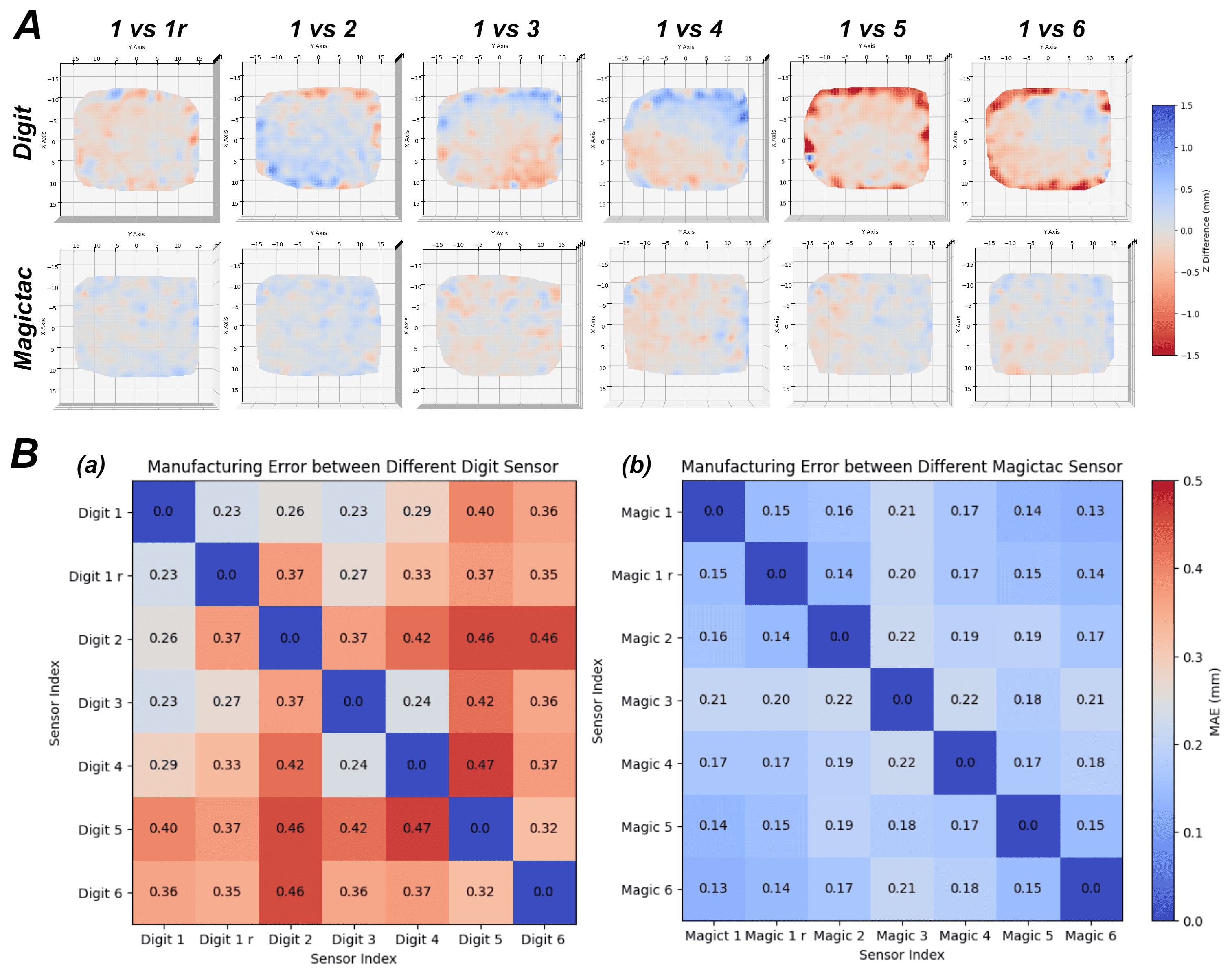}
	\caption{A: Surface error distribution maps for DIGITs (top row) and Mini-MagicTacs (bottom row), where '1r' represents the self-repeat test. B: (a) Manufacturing error matrix between DIGITs; (b) Manufacturing error matrix between Mini-MagicTacs.}
	\label{manufacturing error result}
\end{figure}

To quantitatively analyse such manufacturing error, we decided to reconstruct their surface of each sample. As shown in Fig. \ref{manufacturing error setup} (B), an area with a width and length range of [-12 mm, 12 mm] and [-16 mm, 16 mm] was selected to cover the entire sensor surface. We then randomly sampled this area using the same setup as in the force estimation task, with the only difference being that a threshold was set for the normal force to determine contact. Based on 400 collected points (Fig. \ref{manufacturing error setup} (B.a)), the final sensor surface reconstruction was obtained after noise filtering and curve surface fitting, as shown in Fig. \ref{manufacturing error setup} (B.b). The comparison results of the error distribution map are summarised in Fig. \ref{manufacturing error result} (A). Among six DIGITs, the distributed error between samples 1 and 2 is small, but samples 3 and 4 exhibit a significant skew in the upper right corner compared to sample 1, while samples 5 and 6 are noticeably thicker overall. In contrast, Mini-MagicTacs show consistent differences between units. Then, their global errors are illustrated in Fig. \ref{manufacturing error result} (B). Both DIGITs and Mini-MagicTacs show a self-repeat error of around 0.2 mm due to the baseline error from reconstruction algorithm. However, once extended to the other five samples, the global error of DIGITs falls between 0.3 mm and 0.4 mm, while Mini-MagicTacs only range from 0.15 mm to 0.2 mm. Such results demonstrate that integral printing technology can ensure product quality of Mini-MagicTac compared to the manufacturing method of SOTA VBTS.

\subsubsection{Mechanical Robustness}

\begin{figure}[!htbp]
	\centering
	\includegraphics[width = 1\hsize]{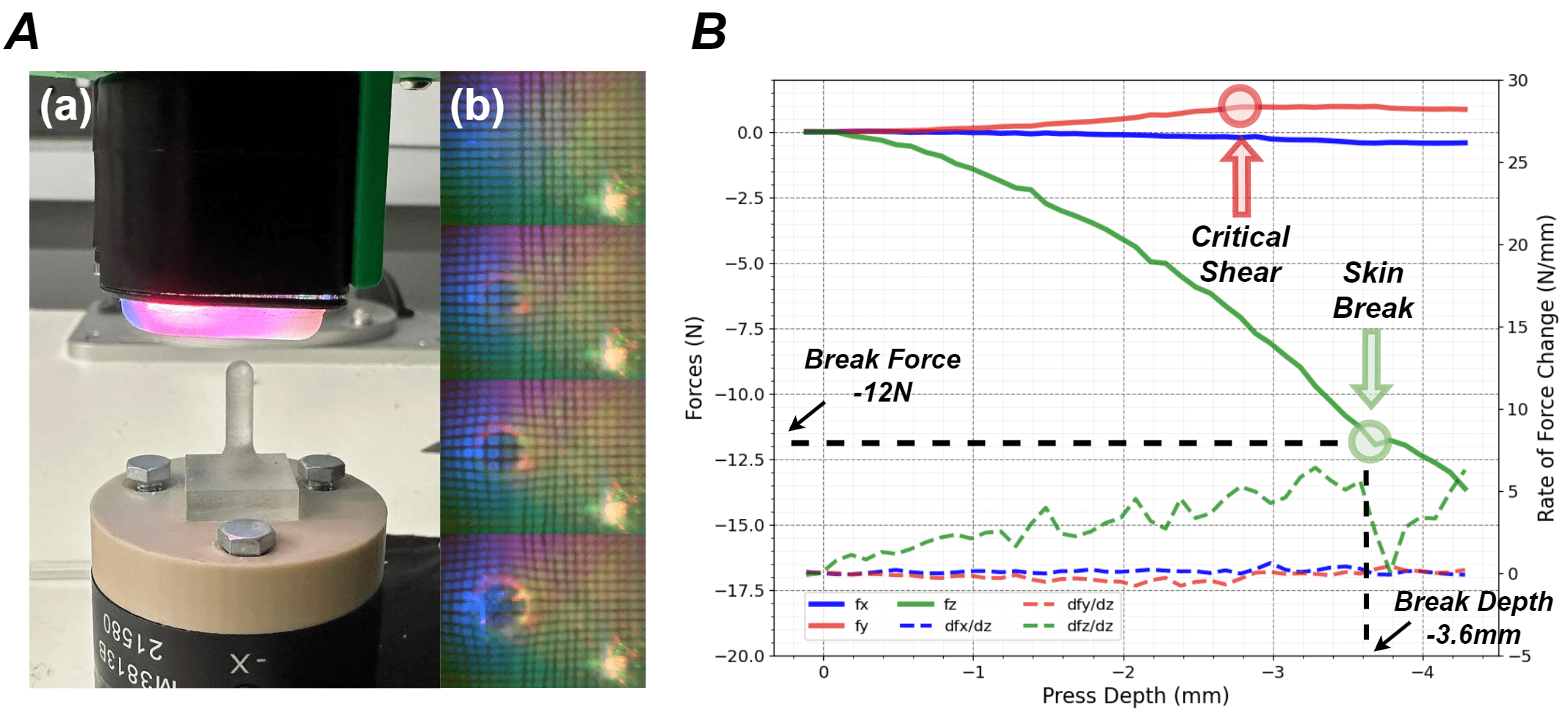}
	\caption{A: (a) Mechanical robustness evaluation through destructive skin puncture testing; (b) The entire process when indentor piercing the skin. B: Relationship analysis between contact force and press depth during destructive skin puncture testing.}
	\label{huge force press setup}
\end{figure}

Mechanical robustness of Mini-MagicTac is closely related to the stability of MagicGripper in practical use, which is primarily influenced by two factors: the thickness of the outer skin and the pressure exerted on the support material within the multi-layer grid. As shown in Fig. \ref{elastomer_stiffness}, the skin thickness affects the mechanical property of the printed elastomer. For instance, thicker skins offer greater resistance to external compression compared to thinner design, but also less sensitive. Also, the support material, which fills the inner part of the multi-layer grid, is a crucial component of the printed elastomer; however, it is inherently less strong than the outer skin made of Agilus30 Clear. To evaluate the mechanical robustness of the hybrid grid structure, we introduce a skin puncture experiment. As depicted in Fig. \ref{huge force press setup} (A.a), a domed cylindrical indentor was used with similar setup in force regression task (Fig. \ref{regression} (A)). During the test, the indentor first makes contact with the skin, and as the press depth increases, the multi-layer grid underneath begins to deform. When the press reaches a certain depth, the support material within the grid cell under the contact area starts to undergo plastic deformation. If the pressing still continues, the skin eventually tears due to the inability to sustain it, which can be also seen in Fig. \ref{huge force press setup} (A.b), with the white mark in the lower right corner indicating the effect of support material fracture after skin piercing.

\begin{figure}[!htbp]
	\centering
	\includegraphics[width = 0.9\hsize]{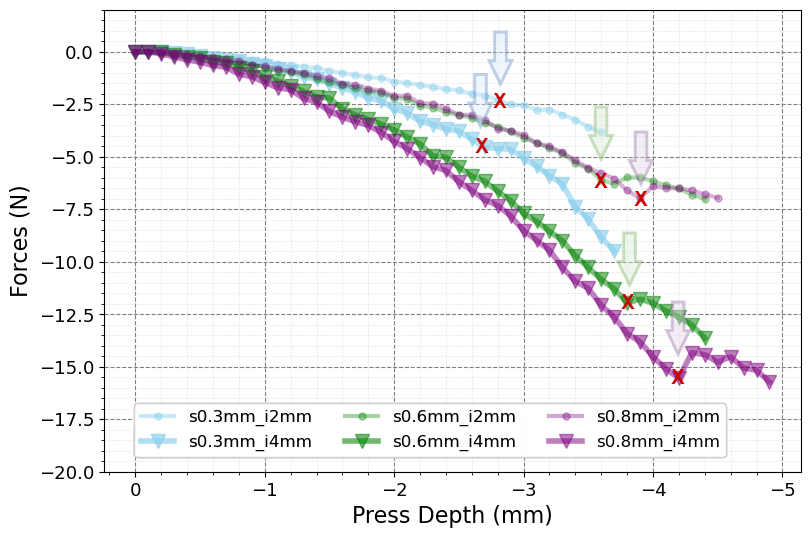}
	\caption{Skin puncture test results with varying skin thicknesses and indentor sizes. The red cross indicates the skin rupture point, and `sXmm\_iYmm' refers to the skin thickness (`Xmm') and indentor diameter size (`Ymm').}
	\label{skin indentor summary}
\end{figure}

This process can be quantitatively analyzed in Fig. \ref{huge force press setup} (B), where Fx, Fy, and Fz are readings from the force sensor beneath the indentor. When Fx or Fy reaches a critical shear force, indicating an inflection point from an increasing trend to a steady state, it suggests the multi-layer grid transitioning from elastic to plastic deformation. At this point, it is unable to convert the normal force into shear force through grid deformation, indicating the limit of the elastic deformation region. Then, Fz at this point can be recorded as a warning threshold to prevent the support structure from breaking, since further increasing Fz may cause its plastic fracture, ultimately resulting in skin rupture (Fig. \ref{huge force press setup} (A.b)). 

For comprehensive evaluation, we used two indentors with diameters of 2 mm and 4 mm, and simultaneously fabricated three Mini-MagicTacs with skin thicknesses of 0.3 mm, 0.6 mm, and 0.8 mm, respectively.  The results of their skin destructive tests are shown in Fig. \ref{skin indentor summary}. It is evident that Mini-MagicTac is more resistant to the larger indenter (4 mm) when the skin thickness keeps same, because the smaller indenter (2 mm) creates greater localized pressure, making the skin more prone to puncturing. Additionally, as the skin thickness increases, Mini-MagicTac can withstand greater maximum pressure and depth. However, as previously mentioned in Fig. \ref{huge force press setup} (B), irreversible damage to the internal support may occur before the skin itself breaks, which necessitates considering changes in the critical shear force. As summarised in Table \ref{mechanical robustness}, the value of Fz at the onset of critical shear is generally lower than the breaking force, supporting the earlier discussion. With skin thickness increasing, the ratio of Fz at critical shear to the breaking force decreases—from 80\% at 0.3 mm skin, to 60\% at 0.6 mm skin, and finally to 50\% at 0.8 mm skin. This is due to the increased mechanical strength from the thicker skin, while the internal support's strength remains largely unchanged. Therefore, it is recommended that the maximum pressing force on Mini-MagicTac should not exceed 6 N, and the pressing depth should not exceed 2.5 mm. In cases of contact with sharp objects, these limits should be constrained to 3 N and 2 mm to ensure the stability of MagicGripper in terms of Mini-MagicTac's mechanical robustness.

\begin{table}[!htbp]
\renewcommand{\arraystretch}{1.5}
\caption{Mechanical Robustness Test of Mini-MagicTac with Different Skin Thickness and Indentor Size}
\centering
\begin{tabular}{cccc}
\hline
\textbf{Sensor}                          & \textbf{Critical Shear} & \textbf{Break Force} & \textbf{Break Depth} \\ \hline
\textit{\textbf{s0.3mm\_i2mm}} & -2.0N  & -2.25N                    & -2.8mm                                              \\
\textit{\textbf{s0.3mm\_i4mm}} & -3.5N  & -4.5N                     & -2.7mm                                              \\
\textit{\textbf{s0.6mm\_i2mm}} & -3.7N  & -6.25N                    & -3.6mm                                              \\
\textit{\textbf{s0.6mm\_i4mm}} & -7.0N  & -12.0N                    & -3.8mm                                              \\
\textit{\textbf{s0.8mm\_i2mm}} & -3.7N  & -7.0N                     & -3.9mm                                              \\
\textit{\textbf{s0.8mm\_i4mm}} & -7.2N  & -15.5N                    & -4.4mm                                              \\ \hline
\end{tabular}
\label{mechanical robustness}
\footnotesize \textbf{Note:} The sensor identifier `sXmm\_iYmm' refers to the skin thickness (`Xmm') and indentor diameter size (`Ymm').
\end{table}

\subsubsection{Performance Robustness}

Unlike visual sensors, tactile sensors are in frequent contact during use, making them susceptible to wear and tear, which can degrade their sensing performance over time. Moreover, the generalisation of Mini-MagicTacs' tactile sensing algorithm in MagicGripper, has yet to be verified even with minimal manufacturing errors. Therefore, a wear and tear experiment was designed, using the same setup as the force estimation task (Fig.~\ref{regression} (A)). The indentor used in skin puncture test was also utilised to rub surface of Mini-MagicTac over an extended period under intensive contact conditions (Fig. \ref{huge force press setup} (A)). Based on Fig.~\ref{skin indentor summary} and Table~\ref{mechanical robustness}, the indentor was chosen to be 6 mm as a larger diameter is less likely to puncture the skin, making it more suitable for long-term wear testing.

\begin{figure}[!htbp]
	\centering
	\includegraphics[width = 1\hsize]{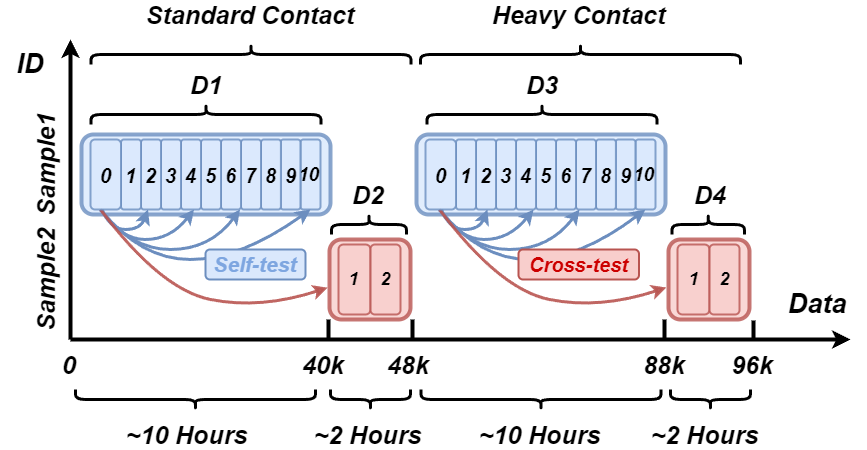}
	\caption{Two Mini-MagicTacs, sample1 and sample2, were used to collect data continuously for 24 hours to verify performance robustness. Four wear and tear datasets - D1, D2, D3, and D4 - were obtained, comprising a total of 96,000 data points. D1 and D3 were divided into 11 segments in chronological order for self-testing, while D2 and D4 were split into halves for cross-testing.}
	\label{performance data collection}
\end{figure}

To simulate worst situations, two identically shaped Mini-MagicTacs with a 0.3 mm thick skin - representing its weaknest version - were selected for performance robustness test. As shown in Fig.~\ref{performance data collection}, two datasets were collected for each sample: the first set involved normal and shear forces controlled within a small range to simulate standard contact wear, while the second set involved larger force ranges to simulate heavy contact wear, also with different amounts of data to simulate real-life scenarios. In details, four datasets were collected in total: D1, D2, D3, and D4. D1 and D3 were collected from Mini-MagicTac sample 1, each containing 39,815 images - close to 40,000. Meanwhile, D2 and D4 were collected from sample 2, each containing 7,945 images - nearly 8,000. The difference between the standard contact data (D1, D2) and heavy contact data (D3, D4) lies in the press depth (1.2 mm vs. 1.6 mm) and shear range (1 mm vs. 1.5 mm), which resulted in different normal/shear force ranges: [(2 N, 2 N, 2 N) vs. (3 N, 3 N, 4 N)].

\begin{figure}[!htbp]
	\centering
	\includegraphics[width = 1\hsize]{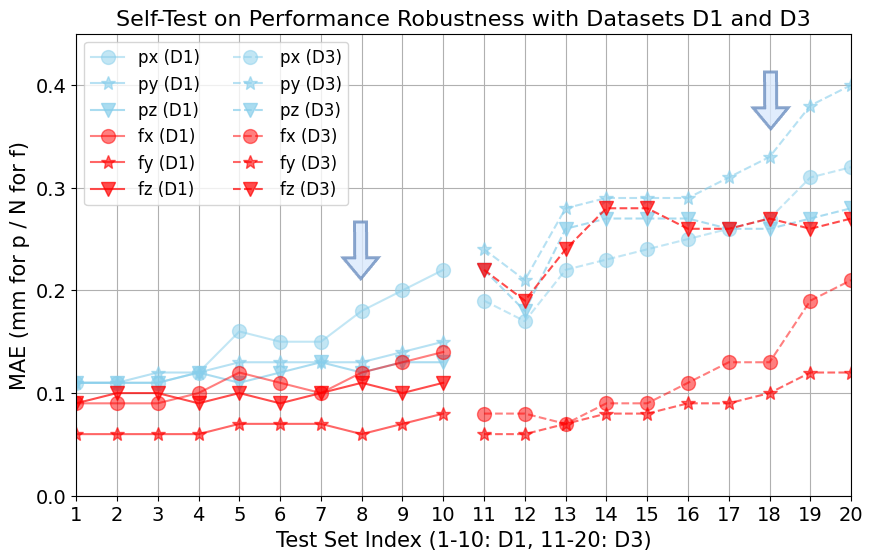}
	\caption{Self-test results using sample 1 whose performance progressively declines as wear from continuous contact increases. In the standard contact scenario represented by D1, performance remains stable up to test set 8 (32,000 data points). A similar trend is observed for D3, with stability maintained until test set 18 (32,000 data points).}
	\label{performance self test}
\end{figure}

For the self-test, the first 10\% of data from D1 and D3 (about 4000) was used for initial model pretraining, with the same software setup as in the force regression task (Fig.~\ref{regression}). The remaining 90\% data was divided into 10 test sets (3600 per each) in chronological order and evaluated by the pretrained model, with wear information implicitly embedded in the temporal data order. The self-test results are summarised in Fig. \ref{performance self test}, which indicates that both contact localization and force regression accuracy decreases as the amount of collected data increases for both D1 and D3, confirming that wear and tear occurred in Mini-MagicTac. The rate of deterioration shows that performance declining gently over the first 80\% ($\leq$ 32000), followed by a more pronounced decrease thereafter. Additionally, the sensing performance of D3 was significantly lower than D1, with the prediction mean absolute error (MAE) being roughly twice in D1, except for Fx and Fy. This difference can be attributed to two main reasons: first, during D3 data collection, sample 1 had already undergone the wear and tear from the D1 acquisition, resulting in an initial performance deterioration bias. Second, D3 was collected using a heavy contact method. Therefore, it is advisable not to exceed 32,000 interactions or apply excessive force during contact experiment, which lead to increased wear and tear.

\begin{figure}[!htbp]
	\centering
	\includegraphics[width = 1\hsize]{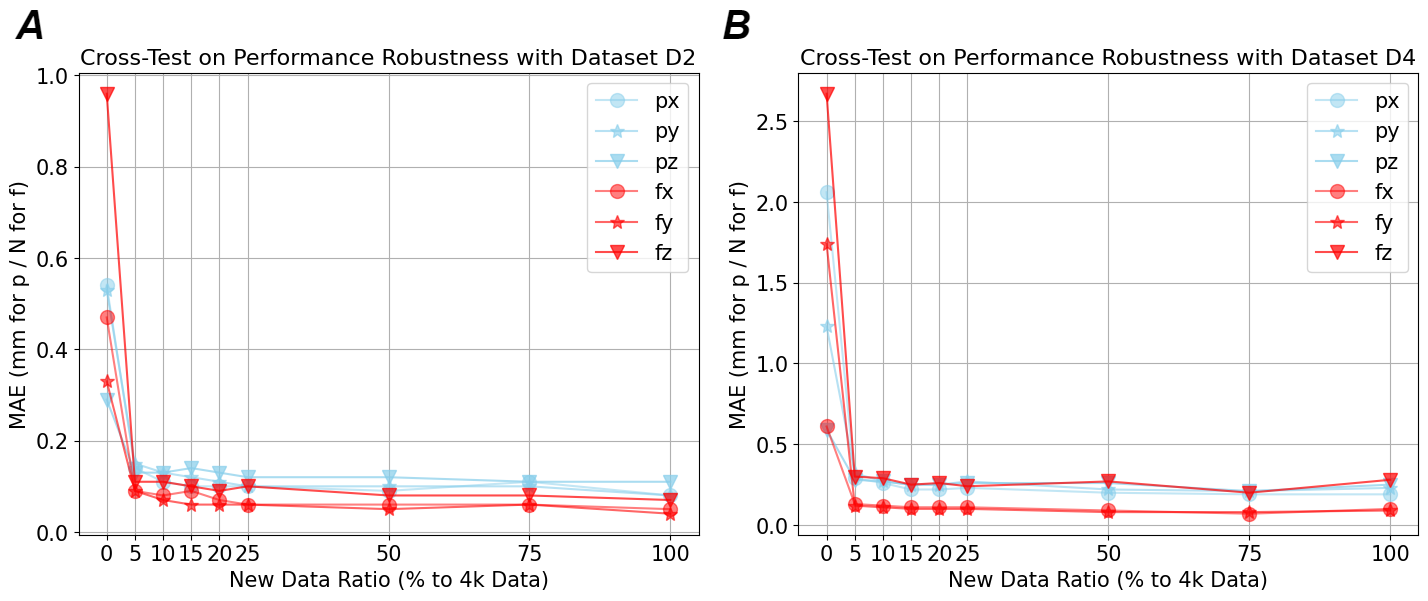}
	\caption{Cross-test results using sample 2 indicate that zero-shot model transfer gap between different Mini-MagicTacs can be reduced by fine-tuning the model with a small amount of new data (5\%), which is effective in both standard contact scenarios (A) and heavy contact scenarios (B).}
	\label{performance cross test}
\end{figure}

\begin{figure*}[]
    \centering
    \includegraphics[width = 0.8\hsize]{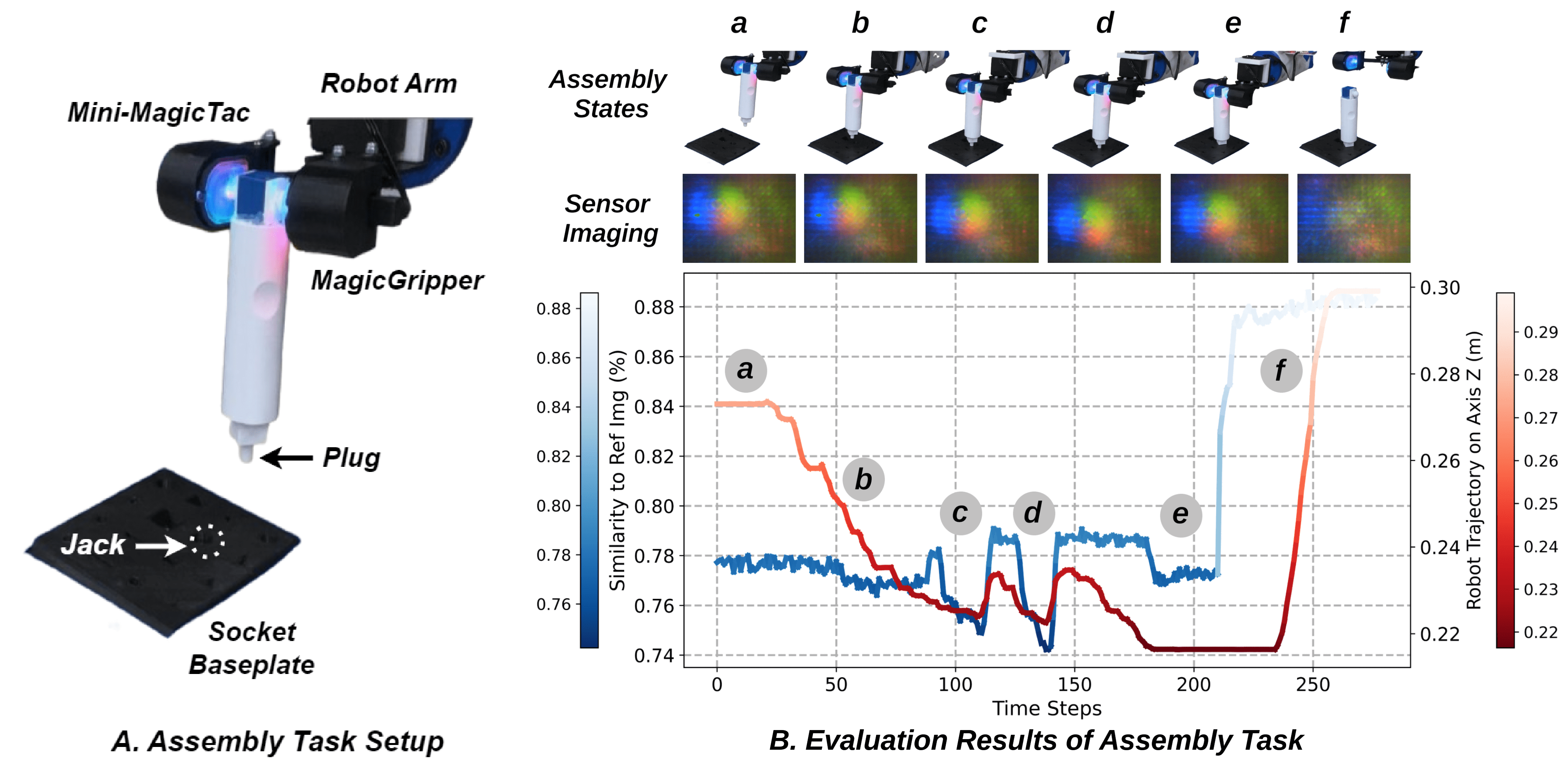}
    \caption{A: Experiment setup of teleoperated assembly task, aiming to assist user accurately insert the plug into the tiny jack on the socket baseplate. B: The teleoperated assembly process with multi-modality sensing of MagicGripper. (a) initial pose, (b) approaching baseplate, (c) first insert trying (failed), (d) second insert trying (failed), (e) third insert trying (succeeded), and (f) end pose.}
    \label{assemble test}
\end{figure*}

For the cross-test, the last 50\% of data from D2 and D4 (about 4000) was used for testing, while the first 50\% was sequentially extracted in increments of 5\%, 10\%, 15\%, 20\%, 25\%, 50\%, 75\%, and 100\% in the temporal order for the fine-tuning task. Those extracted data was then divided into a training set and a validation set in an 8:2 ratio. For example, if 5\% data contained 200 samples, 160 were used for training and remaining 40 for validation. In terms to cross-test, the pretrained model was derived from D1 and D3 of sample 1. The cross-test results are summarised in Fig.~\ref{performance cross test}, which indicates that directly applying data from sample 2 to the pretrained model from sample 1 results in initial errors, suggesting that even with the small manufacturing error, the challenge of zero-shot learning remains. However, by adding a tiny amount of new data - even as little as 5\% (160 data samples for retraining) - the generalisation of the finetuned model to sample 2 can be significantly improved. For instance, the regression MAE after fine-tuning is rapidly reduced to around 0.1 error in D2 and 0.3 error in D4, matching to Fig.~\ref{regression}. It demonstrates that integral manufacturing not only ensures minimal hardware manufacturing error for Mini-MagicTac while also supporting its performance robustness, thereby leading to software generalisability in two-finger MagicGripper.

After above performance and robustness evaluations, Mini-MagicTac demonstrates that it has sufficient capacity to provide reliable tactile sensing, so two Mini-MagicTacs were mounted on MagicGripper to complete the subsequent experiments. The first is a robot assembly task based on teleoperation, another is robot alignment task based on contact pose, and the last is a robot grasp task based on automation.

\subsection{Teleoperated Assembly Task with MagicGripper}

\subsubsection{Experiment Design}

The primary goal of teleoperated assembly task is assisting user to insert the plug into the hole on the base accurately by manipulating the MagicGripper. It is designed to mimic real-world assembly challenges, emphasizing the need for real-time tactile feedback in teleoperation robotic systems. Now make the following assumptions, if the MagicGripper does not align the plug accurately with the hole, the plug will collide with the base, resulting in significantly increased resistance, which acts as a indicator for users to adjust their approach.
In contrast, a successful assembly is indicated by minimal resistance, as the plug fits seamlessly into the jack. The real experiment setup of teleopertation assembly task is illustrated in Fig.~\ref{assemble test} (A), where the black base introduces challenge towards assembly, since it is hard for user to accurately identify whether plug has been socket into the jack through visual-guided teleoperation. For the robotic system, the MagicGripper was employed as the end-effector of the Kinova Gen3 robotic arm. During the experiments, sensor imaging of MagicGripper was conveyed through additional visual cues displayed on a screen, alerting the user to make necessary adjustments. If the peg was successfully inserted into the socket hole, the trial was deemed successful.

Based on Fig.~\ref{multi modality},  multi-modality sensing offers MagicGripper capability to complete the assembly task, first of all, its visual modality can help the user to clearly localize and adjust grasp pose. If the grip position is too close to the gripper edge or sliding, then it can be adjusted by the proximity feature and vision feature. After, the tactile modality provided by the multi-layer grid is also critical, as the degree of structural deformation reflects both static and dynamic tactile features, such as excessive shear forces caused by incorrect mounting positions. In order to alleviate the user's teleoperation workload, we introduce the SSIM (Structural Similarity Index Measurement) metric to assist, which measures the correlation between the MagicGripper's grip states to help users determine whether the assembly task is successfully completed or not.


\subsubsection{Results Analysis}

The teleoperation process are summarized in Fig.~\ref{assemble test} (B), where the blue line indicates the similarity between the last sensing frame and the current frame of MagicGripper,  and the red line represents the robot trajectory along the Z axis. The whole assembly procedure could be divided into 6 sequential stages, as shown in Fig.~\ref{assemble test} (B) (a-f):

\begin{itemize}
\item (a) Initial Positioning: The task commences with the gripper positioned diagonally above the target plate. The user can simultaneously observe the grip state of the gripper and the geometric position of the gripper relative to the assembly point. The sensor image without grasping contact can be set as SSIM reference image.

\item (b) Approach and Alignment: Subsequently, user navigates the gripper approach the plate and carefully adjust its trajectory to align with the jack. During this phase, the Z-axis value of robot arm gradually decreases, bringing the gripper tantalizingly close to the plate. However, due to line-of-sight occlusion, user is difficult to determine whether there is contact or not, while the grid distribution in the sensor image, which remains stationary with high SSIM value (about 0.78), suggests there is no collision.

\item (c) First Insertion Attempt: The initial insertion attempt is marked by a misalignment, resulting in the plug directly contacting the rigid base. This incorrect alignment generates notable shear forces on the MagicGripper's contact surface, leading to deformation of multi-layer grid and decreased SSIM value (0.75). 

\item (d) Second Insertion Attempt: A subsequent adjustment leads to a second insertion attempt from user. Despite undergone the adjustments, this attempt also culminates in a failed assembly, characterized by more severe collisions than the first attempt with even less SSIM value (0.74).

\item (e) Successful Assembly: Learning from the previous attempts, user involves further position adjustment in the third insertion, which ultimately leads to success, as evidenced by the higher SSIM value (0.77).

\item (f) End of Task:  After successful assembly, the gripper opens and transitions to its end pose (f). Also, the grasped object disappeared from its visual sensing area, leading to the hightest SSIM value (0.88).
\end{itemize}


\begin{table}[!htbp]\centering
\renewcommand{\arraystretch}{1.4}
\caption{Identification Accuracy of Misalignment in Teleoperated Assembly Task with and Without MagicGripper}
\begin{tabular}{cc}
\hline
 & \textbf{Identification Accuracy of Misalignment}  \\ \hline
\textbf{With MagicGripper}    & 100\%    \\ 
\textbf{Without MagicGripper}    & 25\%     \\ \hline
\end{tabular}
\label{hole}
\end{table}

In real study, 16 trials of repetitive tests were applied aimed at comparing the effectiveness of assembly task, with and without MagicGripper, in detecting misalignment. Initial misalignments between plug and hole were deliberately introduced by applying slight angular deviations and lateral offsets exceeding 3 mm, simulating common errors during the initial phases of assembly tasks. From Table \ref{hole}, test results demonstrate a performance improvement with MagicGripper, achieving a 100\% success rate in identifying misalignment, compared to a 25\% success rate without it.  This phase exemplifies the user to adapt and correct its teleoperation approach based on MagicGripper's multi-modality sensing, which alleviates the iterative nature of randomized robotic exploration in response to vision-guided feedback.

\begin{figure*}[!htbp]
	\centering
	\includegraphics[width = 1\hsize]{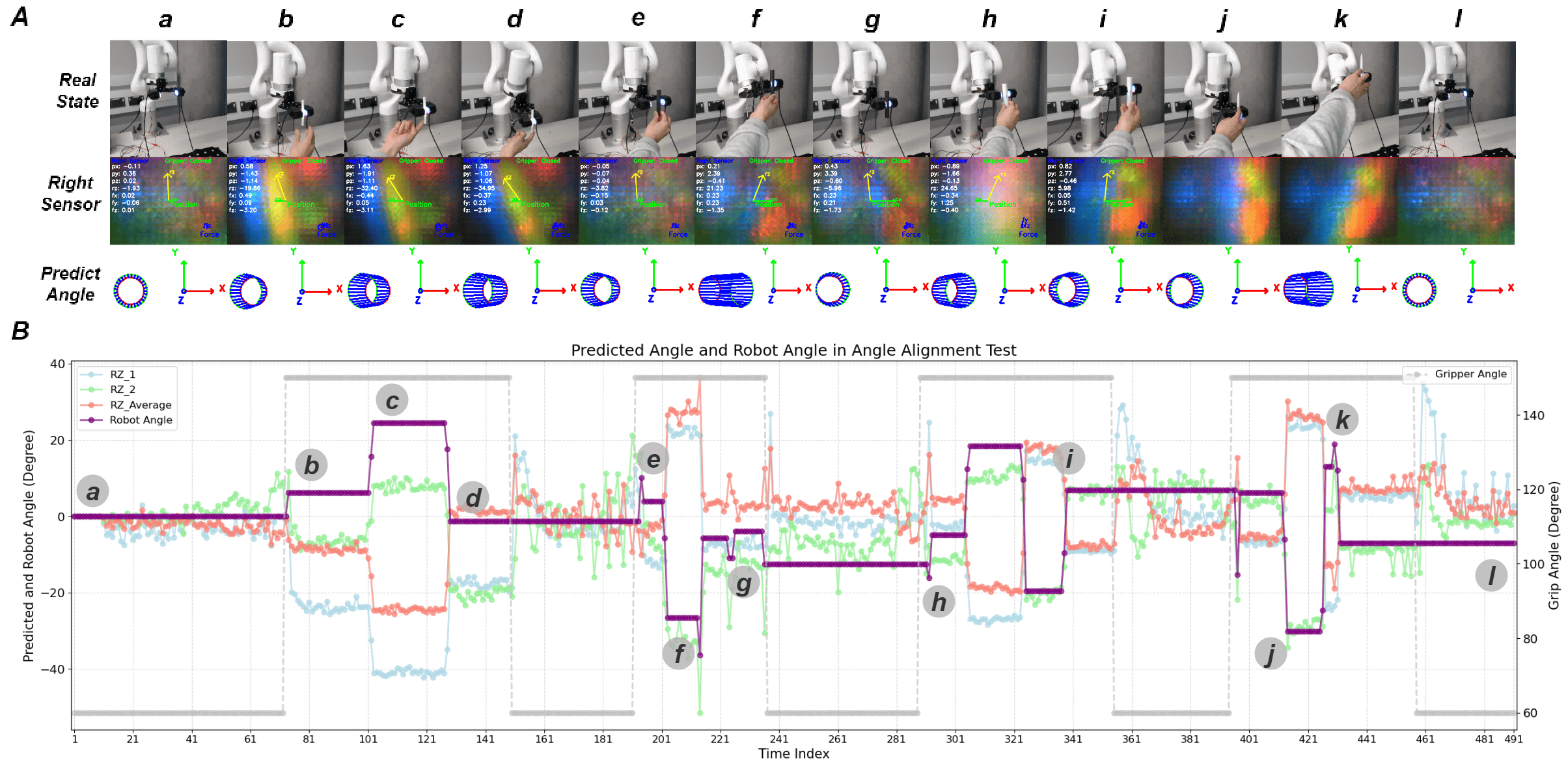}
	\caption{A: Three printing parts of different sizes and colors were selected for testing, along with a pen as an additional test object. B: By combining the angle predictions from both sensors, more accurate real-time angle information can be obtained. The robot arm's rotation angle is set to reverse alignment mode, ensuring that the object to be grasped remains perpendicular to the bottom surface.}
 \vspace{-0.2cm}
	\label{angle alignment}
\end{figure*}

\subsection{Contact Alignment Task with MagicGripper}

While the teleoperated assembly task highlights MagicGripper’s capabilities in tactile-enhanced robotic manipulation, the use of similarity-based tactile inference and stochastic exploration-based control strategies does not fully demonstrate the sensor's potential. To further evaluate its responsiveness, we introduce a contact alignment task that involves both object pose and contact force estimation.


\begin{table}[]
\renewcommand{\arraystretch}{1.5}
\caption{Performance Evaluation using Different Models}
\centering
\begin{tabular}{cccccccc}
\hline
\textbf{Test Error}                        & \textbf{Px} & \textbf{Py} & \textbf{Pz} & \textbf{Fx} & \textbf{Fy} & \textbf{Fz} & \textbf{Rz} \\ \hline
\textit{\textbf{Pretrained Model}}    & 1.12        & 0.16        & 0.06        & 0.07        & 0.06        & 0.12        & 1.84        \\
\textit{\textbf{New (no finetune)}}   & 2.26        & 1.40        & 0.43        & 0.82        & 0.49        & 1.85        & 12.8        \\
\textit{\textbf{New (finetuned)}} & 1.18        & 0.15        & 0.08        & 0.18        & 0.12        & 0.28        & 2.46        \\ \hline
\end{tabular}
\label{model finetune improve}
\end{table}

\subsubsection{Experiment Design}

Firstly,  six cylindrical 3D-printed parts, each 10 cm in length, were fabricated with diameters of 10 mm, 15 mm, and 20 mm, in both black and white.
For the contact alignment task, the MagicGripper needs to sense the rotation angle of the cylindrical parts while clamping them and adjust the robotic arm's angle to ensure the object remains perpendicular to the ground. Before conducting specific experiments, we needed to complete the prediction model training and model transfer between two Mini-MagicTacs in MagicGripper. The model in Fig.~\ref{regression} can not be used directly, as it lacked the ability to predict angles of contact object, necessitating data collection for the six new cylindrical parts.

During this process, three different Mini-MagicTacs were employed to validate the patterns previously analysed in the performance cross-test (Fig. \ref{performance cross test}). First two of Mini-MagicTacs, \textbf{sensor1} and \textbf{sensor2}, were used as the left and right sensors in MagicGripper, while \textbf{sensor3} was used for large-volume data acquisition and model pretraining. Based on sensor3, each print sample was collected 1,800 data, approximately 10,600 in total for 6 samples. 
The collected labels included the relative position of the cylinder's midpoint to sensor center (px, py, pz), the rotation angle of the cylinder's mid-axis relative to the sensor's Y-axis (rz), and the normal force (fz) with shear forces (fx, fy) generated by the cylinder's contact. Such dataset was split in a 7:2:1 ratio to create the training, validation, and test sets, and the same model training setup were used as the force estimation task (Fig.~\ref{regression}), whose test results are summarized in Table \ref{model finetune improve}, referred to as the \textbf{Pretrained Model} (test loss of 0.49).
Next, we used sensor1 and sensor2 to randomly assign three print samples, each collecting 160 data samples, resulting in a total of 960 new data. Half of such dataset (480) was left for testing, while 80\%(384)/20\%(96) of left were used for training/validation to fine-tune the pretrained model. To verify the conclusion from the cross-test (Fig.~\ref{performance cross test}), we compared the model performance before and after fine-tuning, as shown in Table \ref{model finetune improve}. The results show that the model's test performance without fine-tuning (test loss of 2.88) was significantly worse than the fine-tuned model (test loss of 0.64). It demonstrates that using less than 5\% of new Mini-MagicTac data for fine-tuning leads to a substantial reduction in model transfer error. Then, the estimated angles of both sensor1 and sensor2 will be averaged and negatived to adjust the robot pose in terms of object alignment objective.

\subsubsection{Experiment Analysis}

The real experimental setup for the contact alignment task is shown in Fig.~\ref{angle alignment} (A). Three of the six cylindrical parts were randomly selected for the real robot test: a 10 mm white print, a 15 mm black print, and a 20 mm white print. The actual results demonstrate that the fine-tuned model accurately perceives the angular pose of cylinders with varying colors (Fig. \ref{angle alignment} (A.e-g)) and diameters (Fig. \ref{angle alignment} (A.h-i)). To evaluate the generalization of such model, we conducted an additional test on MagicGripper using a pen (Fig. \ref{angle alignment} (A.j-k)), which also yielded positive results, confirming the model's validity.
More detailed experimental data are presented in Fig. \ref{angle alignment} (B), which illustrates the angle prediction from Mini-MagicTac and corresponding robot action of MagicGripper pose. When there is no contact happened (Fig. \ref{angle alignment} (A.a/l)), both the estimated contact force and predict angle is close to zero, only with some noise at task ending (Fig. \ref{angle alignment} (B.a/l)). Once the object has been grabbed with pose changing (Fig. \ref{angle alignment} (A.a-d)), the robot pose of MagicGripper (purple line) can follow such events immediately (Fig. \ref{angle alignment} (B.a-d)). The similar performance has been demonstrated in terms of different object color (Fig. \ref{angle alignment} (B.e-g)) and dimension (Fig. \ref{angle alignment} (B.h-i)), which can even generalize to daily items, such as a pen (Fig. \ref{angle alignment} (B.j-k)).

\begin{figure}[!htbp]
	\centering
	\includegraphics[width = 0.95\hsize]{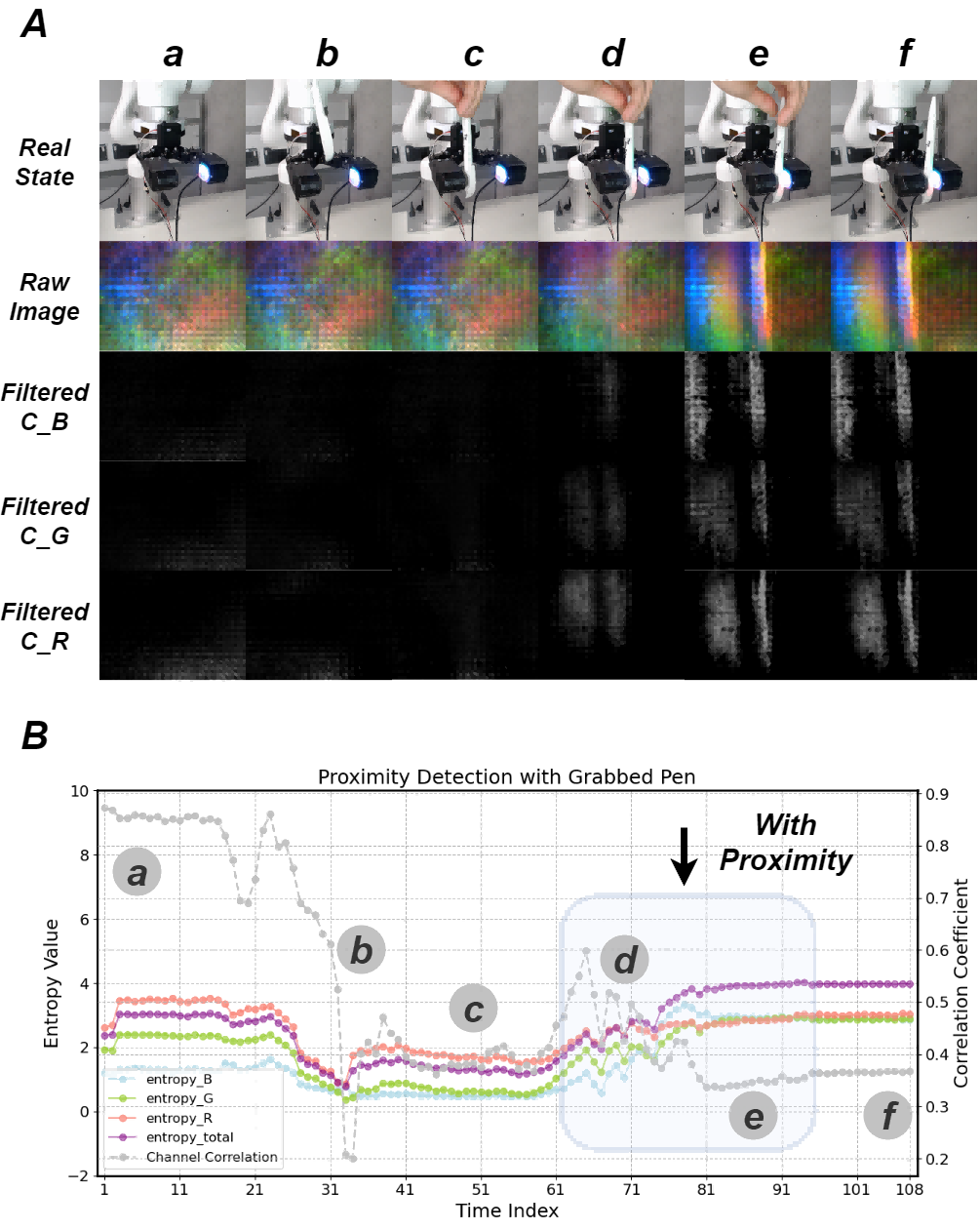}
	\caption{A: The feature distribution across channels deviates when external objects are approaching. B: This deviation results in a decrease in channel correlation coefficient and an increase in entropy due to visual features.}
 \vspace{-0.2cm}
	\label{proximity pen}
\end{figure}

\begin{figure}[!htbp]
	\centering
	\includegraphics[width = 0.95\hsize]{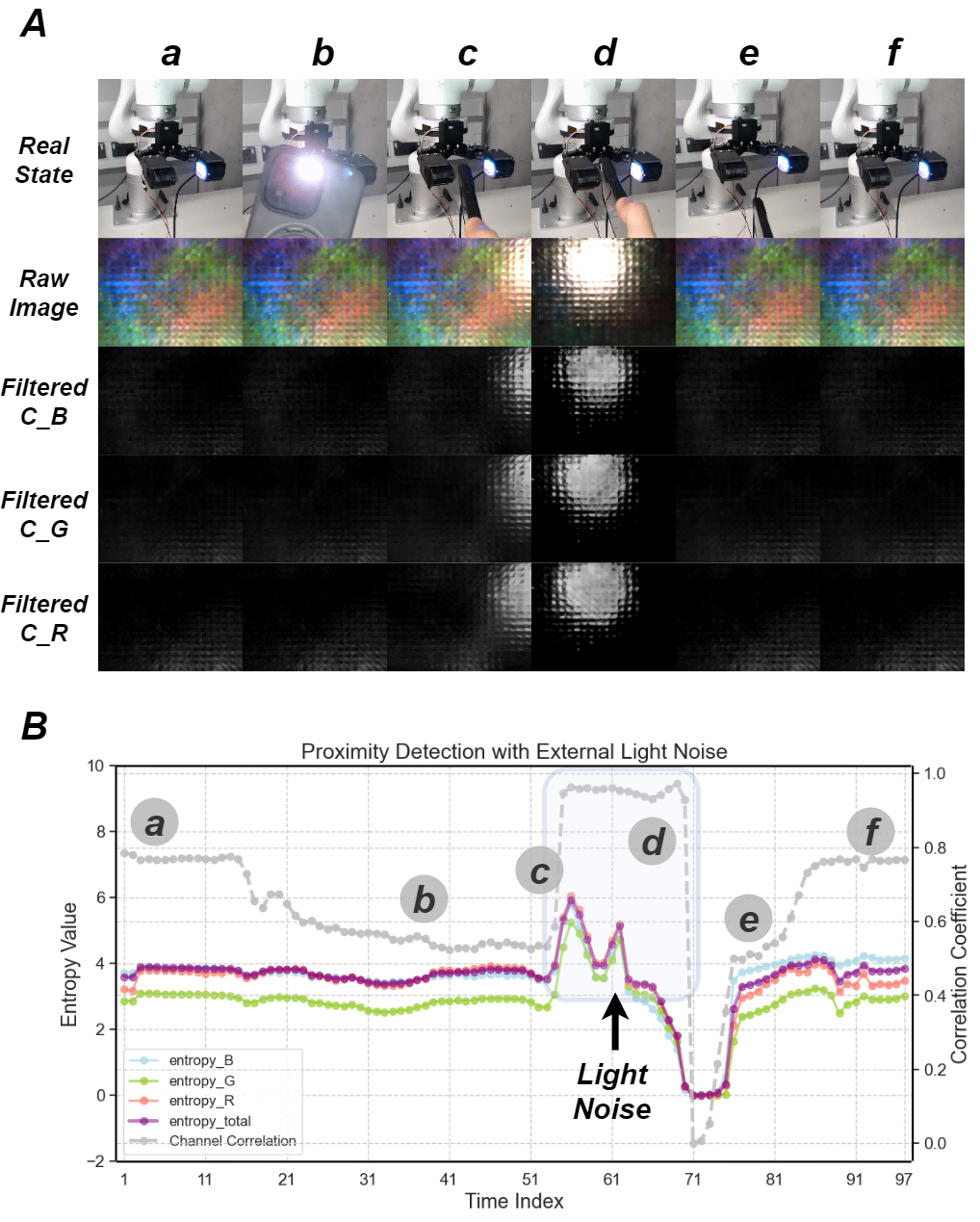}
	\caption{A: External light noise affects all channels uniformly. B: When noise occurs, the correlation coefficients between the channels are high, while the entropy within each channel increases due to changes in lighting environment.}
 \vspace{-0.2cm}
	\label{proximity light noise}
\end{figure}

\begin{figure}[!htbp]
	\centering
	\includegraphics[width = 0.95\hsize]{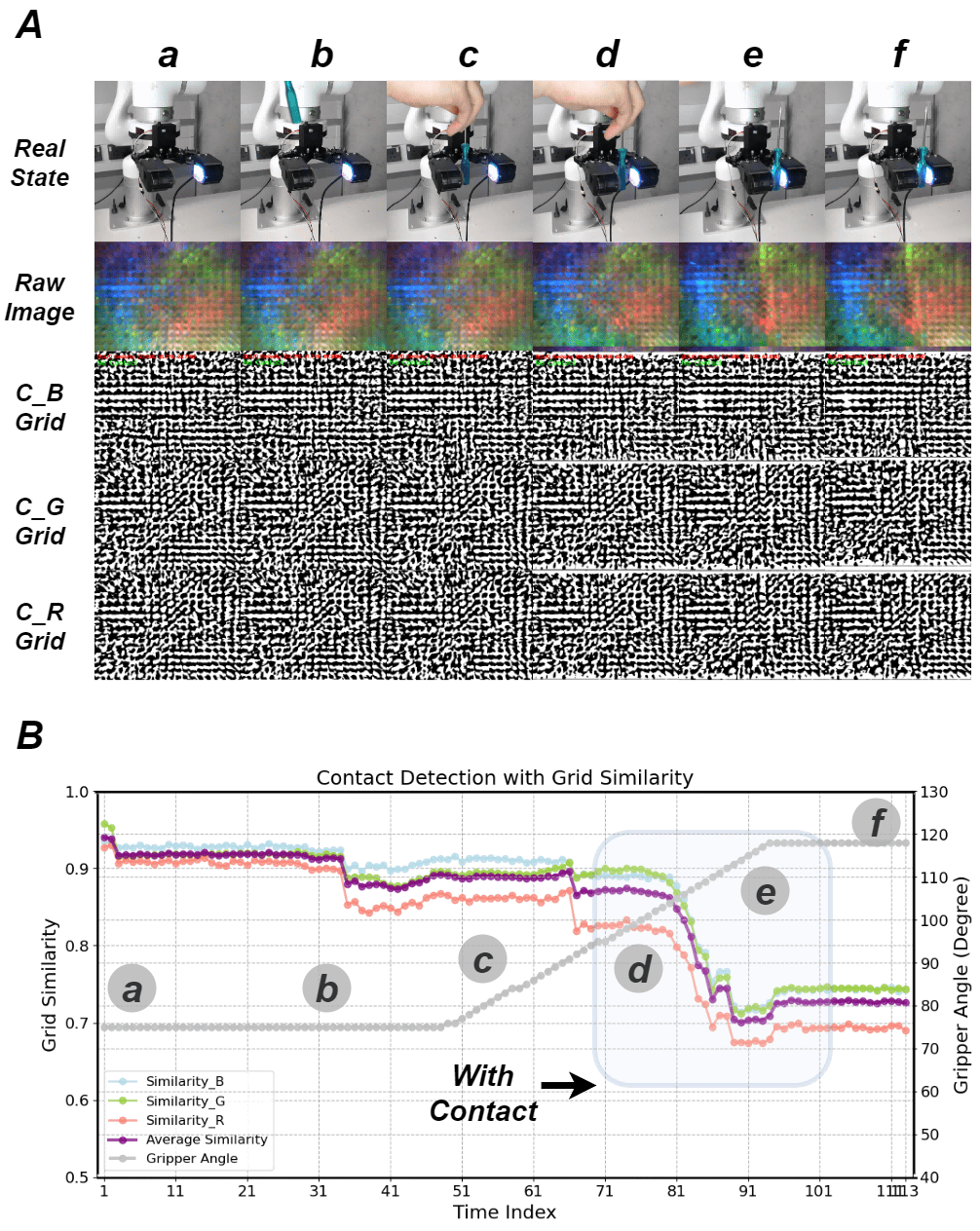}
	\caption{A: The grid structure deforms only after contact occurs. Its geometries between channels vary, but exhibit a consistent trend upon contact. B: As MagicGripper contacts with the screwdriver, the grid similarity across channels significantly decreases.}
 \vspace{-0.2cm}
	\label{contact screw}
\end{figure}

\begin{figure*}[!htbp]
	\centering
	\includegraphics[width = 1\hsize]{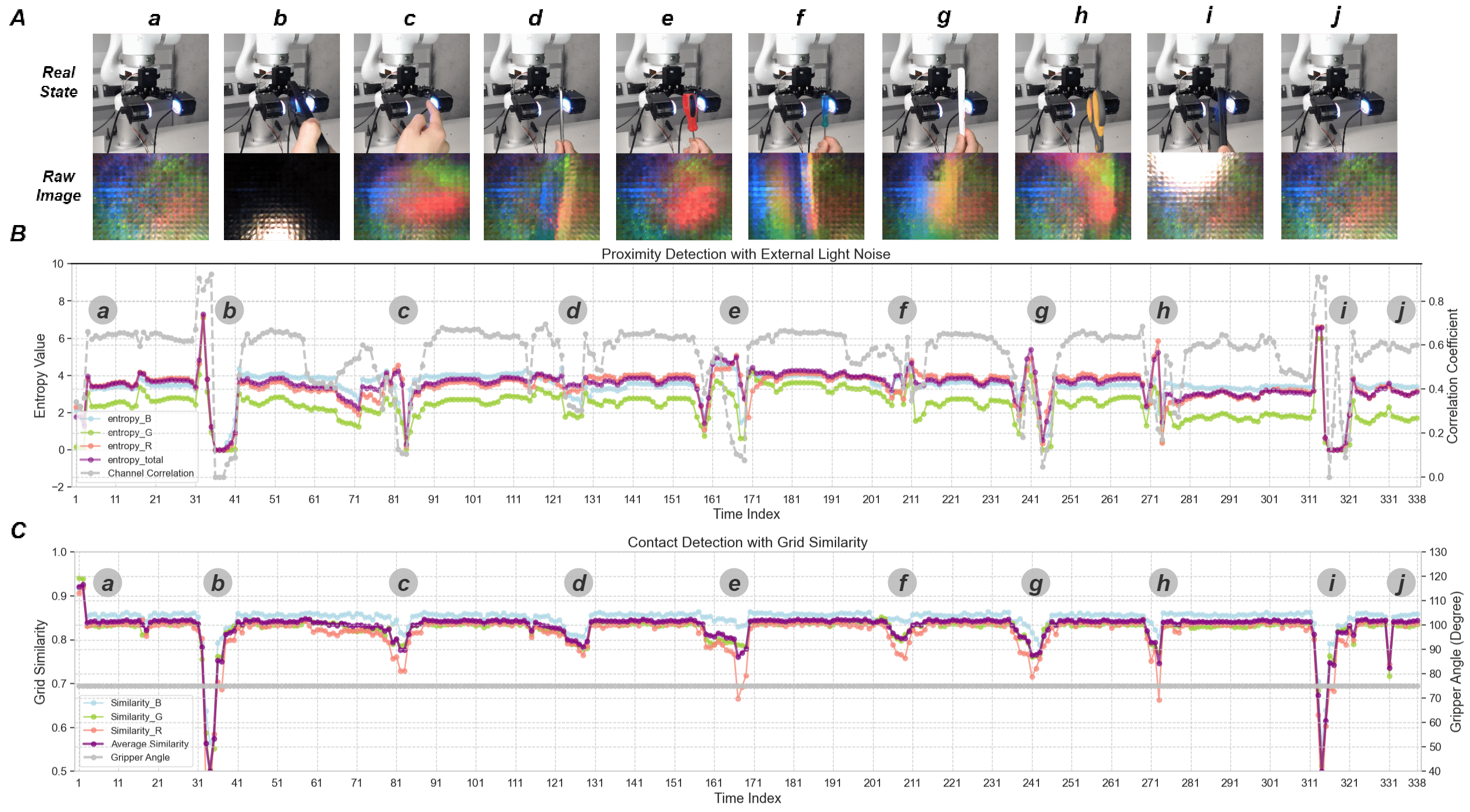}
	\caption{A: Mixed noise, proximity, contact tests on MagicTac, where six different objects are selected for the experiment. B: Significant difference between the effects of light interference and object proximity on channel entropy and correlation coefficient. C: Even with light contact, grid similarity can change, and light noise can have an effect, but it tends to be very dramatic.}
 \vspace{-0.2cm}
	\label{proximity contact multi stuff}
\end{figure*}

\subsection{Autonomous Robotic Grasping Test with MagicGripper}

To verify the effectiveness of the proposed proximity and contact detection algorithm illustrated in Fig.~\ref{proximity contact framework}, an autonomous robotic grasping test was conducted.

\subsubsection{Experiment Preparation}

Several core functions in algorithm need to be verified before formal experiment, where a flashlight, a pen, and a screwdriver were utilized:

\begin{itemize}
  \item Proximity Detection: Proximity detection function should response to approaching event and has ability to identify between valid event and noise influence. As seen in Fig.~\ref{proximity pen} (A.c/d/e/f), the approaching pen lead to different pattern distribution of the three channels (B, G, R), whose intensity increased as it get closer to MagicGripper. Also as demonstrated in Fig.~\ref{proximity pen} (B.d/e), the channel entropy arise higher while their correlation goes down, which indicates the proximity state as '\textbf{Approaching}' (Algorithm \ref{alg:proximity_detection}). As shown in Fig.~\ref{proximity light noise} (A.c/d), when the flashlight was present, the filtered images showed a high degree of similarity across channels. It led to a sudden increase in the correlation coefficients as illustrated in Fig. \ref{proximity light noise} (B.c/d), even through channel entropy get fluctuated increasing, which represents the proximity state as '\textbf{Noise}' (Algorithm \ref{alg:proximity_detection}).

  \item Contact Detection: Contact detection function aims to provide touch probability without directly tracking grid itself. As shown in Fig. \ref{contact screw} (A.a/b/c), it can be observed while grid geometry differ across channels, their similarity to the grid reference mask remains stable when no contact happened, which is attributed to the maximum probability distribution achieved through temporal fusion (Algorithm \ref{alg:TF function}). As MagicGripper close from open (gray line), the object (screwdriver) gradually made contact with Mini-MagicTac, causing grid similarity to decrease (Fig. \ref{contact screw} (B.d/e)), which indicates the contact state as '\textbf{Touched}' (Algorithm \ref{alg:contact_detection}).

  \item Combined Detection: To further verify the algorithm's generalisation, we conducted a combined proximity and contact detection test using six common items from daily life (Fig. \ref{proximity contact multi stuff} (A)): a human finger (c), a silver metal plate hand (d), a red rubber screwdriver (e), a green plastic screwdriver (f), a white plastic pen (g), and yellow rubber scissors (h). Random light noise was added at the beginning (b) and end of the experiment (i). Proximity detection results are shown in Fig. \ref{proximity contact multi stuff} (B), where channel entropy fluctuated whenever an object approached or noise was present, with peaks varying according to the characteristics of the object. For instance, the white pen (g) and yellow scissors (h) caused greater entropy changes compared to the silver spanner (d), since metal reflected light in a particular direction, leading to less information being captured. In contrast, the correlation coefficient exhibited a more steady decreasing when an object approached and directly increasing to 1 when light noise was present.
  Then, contact detection results are shown in Fig. \ref{proximity contact multi stuff}(C). Since the MagicGripper was not closed during the test, all object contacts with Mini-MagicTac were made manually. Nonetheless, all touch events were successfully characterized by changes in grid similarity, which dropped from a steady state of 0.85 to below 0.8. It is worth noting that light noise also affected grid similarity, while drastically reducing it to below 0.5 (Fig. \ref{proximity contact multi stuff} (C.b/i)). Therefore, co-analysis between contact state from Algorithm \ref{alg:contact_detection} with proximity state from Algorithm \ref{alg:proximity_detection} is indeed necessary to improve the contact detection robustness to prevent misclassifications.

\end{itemize}

\begin{figure*}[!htbp]
	\centering
	\includegraphics[width = 1\hsize]{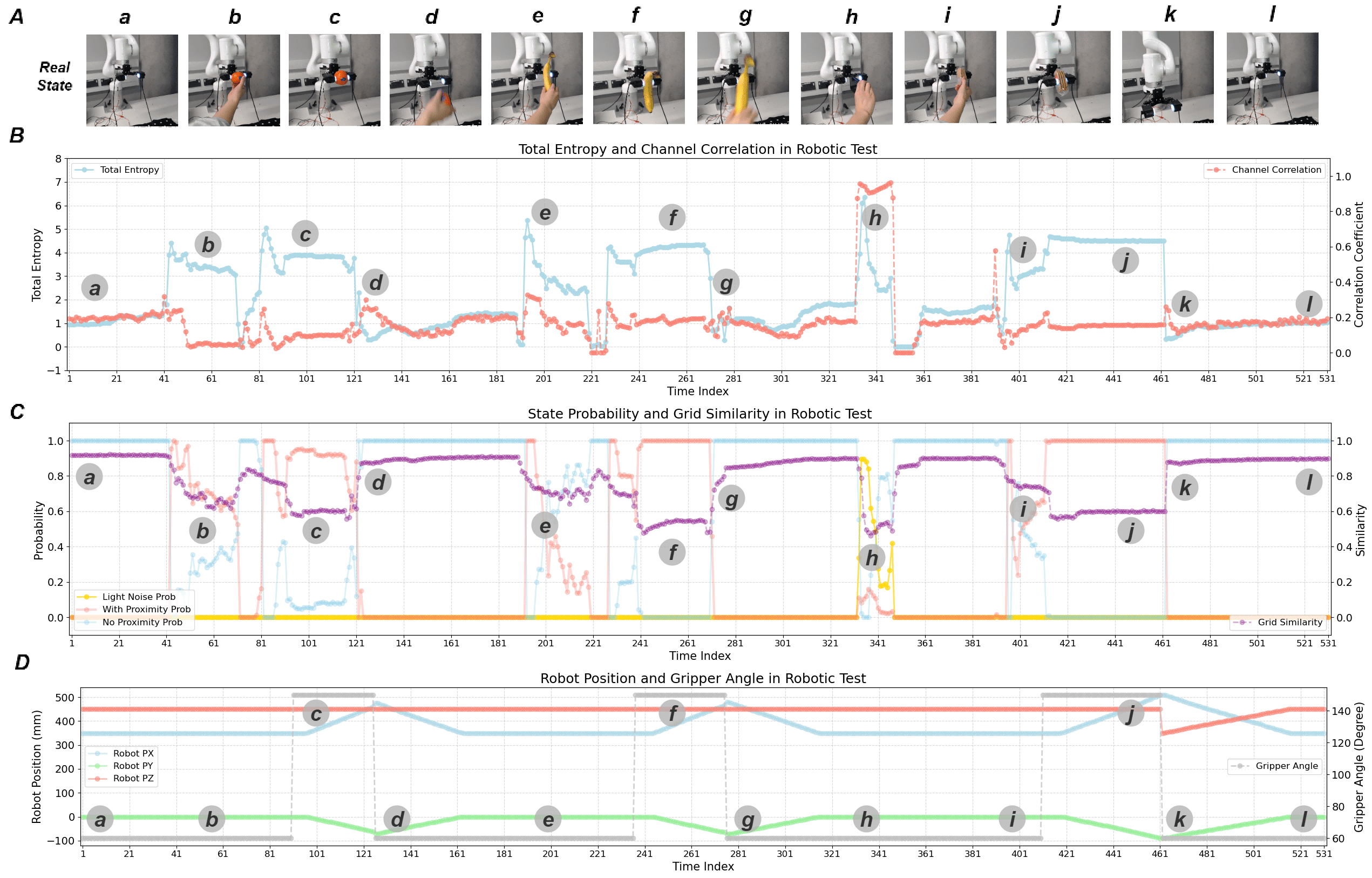}
	\caption{A: In robot grasp test, Magicgripper automatically grabs an object when it approaches, ignoring light noise. After grasping, it transports the object from point A to point B, but returns to point A if object slippage. B: Channel entropy and correlation can distinguish between light noise and approaching objects. C: If object slips, grid similarity can detect such event. D: After two instances of slippage and light noise interference, the robotic arm successfully transported the final object to the target point with MagicGripper.}
 \vspace{-0.2cm}
	\label{robot grab test}
\end{figure*}

\subsubsection{Experiment Design}


Building on above validated algorithms, we designed an autonomous robotic grasping experiment relying on MagicGripper's multi-modality sensing, whose objective is to enable a robot to autonomously perform food transport tasks without human intervention or extra visual sensing (e.g., cameras). In details, MagicGripper was required to: (1) detect and localize the approaching objects, (2) grasp them reliably, (3) monitor for slippage during transportation, and (4) release items at target location. Three everyday food items were selected as test objects: an orange, a banana, and a packet of biscuits, which vary in texture, color, size, weight, and hardness, providing diverse visual and tactile features to challenge the MagicGripper system.

To achieve this goal, proximity detection was firstly used to estimate whether there was an object approaching nearby. The specific probability value was calculated based on the disparities between channel entropy $E_{total}$/correlation coefficient $C_{total}$ and their threshold $\tau_{E}$ and $\tau_{C}$ (Algorithm \ref{alg:proximity_detection}). Once an object was continuously detected as '\textbf{Approaching}', then MagicGripper was activated to grasp it by decreasing gripper angle. At meantime, gripper should be stable when encountering random noise without any invalid grasping attempt. Then contact detection should be activated, if the grid similarity fell below a set threshold $\tau_{G}$, it indicated a successful grasp, allowing the transport operation to commence, where robotic arm moved from the starting point to the target point. During transportation, once grid similarity increased suddenly, it was determined that a slip had occurred, prompting the robotic arm return to the starting point and wait for another proximity event. The experimental setups are shown in Fig. \ref{robot grab test} (A).

\subsubsection{Experiment Analysis}

From Fig. \ref{robot grab test} (B), proximity and contact detection generalizes well to daily items such as fruits. The total entropy $E_{total}$ for orange (b), bananas (e), and biscuits (i) increased to around 4 as they approaching, while their correlation coefficient $C_{total}$ decreased from 0.2 to around 0. Notably, the trends in entropy and correlation coefficient differed from other objects in close proximity, even with light noise present (h). Based on that, the determination of proximity and contact detection is shown in Fig.~\ref{robot grab test} (C). It can be seen that the rise in proximity probability (blue line) tends to synchronize with the decrease in grid similarity (purple line), suggesting a strong correlation between them. If a grid similarity threshold of 0.7 is used for contact detection, proximity is determined 10 to 15 time steps earlier than contact, which was sufficient for the MagicGripper to determine when to trigger grasping (Fig. \ref{robot grab test} (D.c/f/j)). Due to experiment design of intentionally human-created slip event, only the last biscuit grasp successfully delivered to the target location (k). The first two attempts failed because the items were manually pulled away (Fig.~\ref{robot grab test} (A.d/g)), causing the MagicGripper to detect slippage and return to the starting point to wait for a new item to approach ((Fig.~\ref{robot grab test} (C.d/g))). Also, it is worth noting that the entire operation was unaffected by external light noise (Fig.~\ref{robot grab test} (A.h)), where the occurrence of this noise was successfully predicted and accounted for (yellow line) in Fig. \ref{robot grab test} (C.h).

\begin{table*}[!htbp]\centering
\renewcommand{\arraystretch}{1.4}
\caption{Property Comparison of Grid-like and Typical Tactile Sensors}
\begin{tabular}{ccccccc}
\hline

\textbf{Sensor}                 & \textbf{Tactile Mechanism}  & \textbf{Spatial Resolution} & \textbf{Dynamic Sensing} & \textbf{Design Flexibility} & \textbf{Fabrication Cost} & \textbf{Modality Type} \\ \hline
\textit{\textbf{GelSight-type}} & Coating Reflection  & Fine                      & Marker*                            & Low                             & High  & Tactile                 \\
\textit{\textbf{MD-based}}      & Marker Distribution & Rough                     & Sensitive                    & Low                             & High    & Tactile              \\
\textit{\textbf{Grid-like}}     & Grid Distribution   & Good                      & Sensitive                    & High                            & Affordable   & Tactile + Visual             \\ \hline
\end{tabular}
\label{comparison}
\end{table*}

\section{Discussions}
\label{Discuss-Future}
As shown in Table~\ref{comparison}, we present a comparative summary of the key properties of our proposed grid-like sensor alongside two typies of representative VBTSs: a traditional marker displacement (MD)-based sensor and a GelSight-type sensor.

Thanks to uniform distribution of grid cells, the grid-like sensor surpasses the MD-based sensor in static tactile sensing, which typically relies on sparsely distributed markers. Moreover, the embedded 3D grid structure within the elastomer enables superior dynamic tactile sensing compared to the GelSight-type sensor, which often depends on additional markers to extract dynamic tactile features. Additionally, the use of integral printing in its fabrication provides advantages in terms of structural customization and product quality, reducing the need for manual assembly steps common in the other two designs. Finally, the grid-like sensor inherently supports multimodal sensing, integrating visual and tactile modalities, which includes a wide range of sub-modal information, further distinguishing it as a versatile and scalable solution for contact-rich manipulation tasks.

Looking ahead, we aim to extend MagicGripper towards various robotic manipulation scenarios. Targeted applications include autonomous assembly of intricate structures, where fine spatial awareness and robust contact sensing are essential. Moreover, integrating mini-MagicTac into a dexterous five-finger robotic hand could significantly improve tasks requiring precise in-hand manipulation of delicate objects. Beyond industrial use, we also see promising opportunities in medical diagnostics, particularly in tumor detection, where high sensitivity and spatial resolution are critical for identifying abnormal tissue masses.

\section{Conclusions}
\label{Conclusions}

This paper presents a novel gripper, \textit{MagicGripper}, which integrates seamlessly with a grid-structured optical tactile sensor, \textit{mini-MagicTac}. This innovative design offers several key advantages, including high spatial resolution, sensitive contact detection, and robust multi-modal sensing capabilities. Its customizability and affordability are enhanced by multi-material 3D printing, which simplifies fabrication and enables design flexibility. To support robotic manipulation tasks, we propose an algorithmic framework for proximity and contact detection by utilizing channel entropy and correlation. Extensive experiments validate the effectiveness of MagicGripper across various scenarios, which delivers 0.15mm spatial resolution, quarter-millimetre level 3D positioning accuracy, millinewton-level contact force prediction accuracy, less than 0.2mm manufacturing quality, and high-performance fine-tuning with only $5\%$ of the data required. Especially, its multi-modal sensing capability in compact structure can reliably detect and distinguishes contact and proximity states or even light noise, which has represented advantages in robot teleoperation and autonomous robot grasping task.

\bibliographystyle{IEEEtran}
\bibliography{references}

\begin{thebibliography}{10}
\providecommand{\url}[1]{#1}
\csname url@samestyle\endcsname
\providecommand{\newblock}{\relax}
\providecommand{\bibinfo}[2]{#2}
\providecommand{\BIBentrySTDinterwordspacing}{\spaceskip=0pt\relax}
\providecommand{\BIBentryALTinterwordstretchfactor}{4}
\providecommand{\BIBentryALTinterwordspacing}{\spaceskip=\fontdimen2\font plus
\BIBentryALTinterwordstretchfactor\fontdimen3\font minus \fontdimen4\font\relax}
\providecommand{\BIBforeignlanguage}[2]{{%
\expandafter\ifx\csname l@#1\endcsname\relax
\typeout{** WARNING: IEEEtran.bst: No hyphenation pattern has been}%
\typeout{** loaded for the language `#1'. Using the pattern for}%
\typeout{** the default language instead.}%
\else
\language=\csname l@#1\endcsname
\fi
#2}}
\providecommand{\BIBdecl}{\relax}
\BIBdecl

\bibitem{fan2024mag}
W.~Fan, H.~Li, and D.~Zhang, ``Magictac: A novel high-resolution 3d multi-layer grid-based tactile sensor,'' \emph{2024 IEEE International Conference on Robotics and Automation (ICRA)}, 2024.

\bibitem{fan2024crystaltac}
------, ``Crystaltac: Vision-based tactile sensor family fabricated via rapid monolithic manufacturing,'' \emph{Cyborg and Bionic Systems}, 2024.

\bibitem{yuan2017gelsight}
W.~Yuan, S.~Dong, and E.~H. Adelson, ``Gelsight: High-resolution robot tactile sensors for estimating geometry and force,'' \emph{Sensors}, vol.~17, no.~12, p. 2762, 2017.

\bibitem{gomes2020geltip}
D.~F. Gomes, Z.~Lin, and S.~Luo, ``Geltip: A finger-shaped optical tactile sensor for robotic manipulation,'' in \emph{2020 IEEE/RSJ International Conference on Intelligent Robots and Systems (IROS)}.\hskip 1em plus 0.5em minus 0.4em\relax IEEE, 2020, pp. 9903--9909.

\bibitem{lambeta2020digit}
M.~Lambeta, P.-W. Chou, S.~Tian, B.~Yang, B.~Maloon, V.~R. Most, D.~Stroud, R.~Santos, A.~Byagowi, G.~Kammerer \emph{et~al.}, ``Digit: A novel design for a low-cost compact high-resolution tactile sensor with application to in-hand manipulation,'' \emph{IEEE Robotics and Automation Letters}, vol.~5, no.~3, pp. 3838--3845, 2020.

\bibitem{padmanabha2020omnitact}
A.~Padmanabha, F.~Ebert, S.~Tian, R.~Calandra, C.~Finn, and S.~Levine, ``Omnitact: A multi-directional high-resolution touch sensor,'' in \emph{2020 IEEE International Conference on Robotics and Automation (ICRA)}.\hskip 1em plus 0.5em minus 0.4em\relax IEEE, 2020, pp. 618--624.

\bibitem{li2023marker}
M.~Li, T.~Li, and Y.~Jiang, ``Marker displacement method used in vision-based tactile sensors—from 2d to 3d-a review,'' \emph{IEEE Sensors Journal}, 2023.

\bibitem{yang2021enhanced}
Y.~Yang, X.~Wang, Z.~Zhou, J.~Zeng, and H.~Liu, ``An enhanced fingervision for contact spatial surface sensing,'' \emph{IEEE Sensors Journal}, vol.~21, no.~15, pp. 16\,492--16\,502, 2021.

\bibitem{lepora2022digitac}
N.~F. Lepora, Y.~Lin, B.~Money-Coomes, and J.~Lloyd, ``Digitac: A digit-tactip hybrid tactile sensor for comparing low-cost high-resolution robot touch,'' \emph{IEEE Robotics and Automation Letters}, vol.~7, no.~4, pp. 9382--9388, 2022.

\bibitem{sato2009finger}
K.~Sato, K.~Kamiyama, N.~Kawakami, and S.~Tachi, ``Finger-shaped gelforce: sensor for measuring surface traction fields for robotic hand,'' \emph{IEEE Transactions on Haptics}, vol.~3, no.~1, pp. 37--47, 2009.

\bibitem{lin2019sensing}
X.~Lin and M.~Wiertlewski, ``Sensing the frictional state of a robotic skin via subtractive color mixing,'' \emph{IEEE Robotics and Automation Letters}, vol.~4, no.~3, pp. 2386--2392, 2019.

\bibitem{lin2020curvature}
X.~Lin, L.~Willemet, A.~Bailleul, and M.~Wiertlewski, ``Curvature sensing with a spherical tactile sensor using the color-interference of a marker array,'' in \emph{2020 IEEE International Conference on Robotics and Automation (ICRA)}.\hskip 1em plus 0.5em minus 0.4em\relax IEEE, 2020, pp. 603--609.

\bibitem{taylor2022gelslim}
I.~H. Taylor, S.~Dong, and A.~Rodriguez, ``Gelslim 3.0: High-resolution measurement of shape, force and slip in a compact tactile-sensing finger,'' in \emph{2022 International Conference on Robotics and Automation (ICRA)}.\hskip 1em plus 0.5em minus 0.4em\relax IEEE, 2022, pp. 10\,781--10\,787.

\bibitem{do2023densetact}
W.~K. Do, B.~Jurewicz, and M.~Kennedy, ``Densetact 2.0: Optical tactile sensor for shape and force reconstruction,'' in \emph{2023 IEEE International Conference on Robotics and Automation (ICRA)}.\hskip 1em plus 0.5em minus 0.4em\relax IEEE, 2023, pp. 12\,549--12\,555.

\bibitem{zhang2023gelstereo}
C.~Zhang, S.~Cui, S.~Wang, J.~Hu, Y.~Cai, R.~Wang, and Y.~Wang, ``Gelstereo 2.0: An improved gelstereo sensor with multimedium refractive stereo calibration,'' \emph{IEEE Transactions on Industrial Electronics}, 2023.

\bibitem{yamaguchi2016combining}
A.~Yamaguchi and C.~G. Atkeson, ``Combining finger vision and optical tactile sensing: Reducing and handling errors while cutting vegetables,'' in \emph{2016 IEEE-RAS 16th International Conference on Humanoid Robots (Humanoids)}.\hskip 1em plus 0.5em minus 0.4em\relax IEEE, 2016, pp. 1045--1051.

\bibitem{yamaguchi2021fingervision}
A.~Yamaguchi, ``Fingervision with whiskers: Light touch detection with vision-based tactile sensors,'' in \emph{2021 Fifth IEEE International Conference on Robotic Computing (IRC)}.\hskip 1em plus 0.5em minus 0.4em\relax IEEE, 2021, pp. 56--64.

\bibitem{fan2024vitactip}
W.~Fan, H.~Li, W.~Si, S.~Luo, N.~Lepora, and D.~Zhang, ``Vitactip: Design and verification of a novel biomimetic physical vision-tactile fusion sensor,'' in \emph{2024 IEEE International Conference on Robotics and Automation (ICRA)}.\hskip 1em plus 0.5em minus 0.4em\relax IEEE, 2024, pp. 1056--1062.

\bibitem{kim2022uvtac}
W.~Kim, W.~D. Kim, J.-J. Kim, C.-H. Kim, and J.~Kim, ``Uvtac: Switchable uv marker-based tactile sensing finger for effective force estimation and object localization,'' \emph{IEEE Robotics and Automation Letters}, vol.~7, no.~3, pp. 6036--6043, 2022.

\bibitem{wang2022spectac}
Q.~Wang, Y.~Du, and M.~Y. Wang, ``Spectac: A visual-tactile dual-modality sensor using uv illumination,'' in \emph{2022 International Conference on Robotics and Automation (ICRA)}.\hskip 1em plus 0.5em minus 0.4em\relax IEEE, 2022, pp. 10\,844--10\,850.

\bibitem{hogan2021seeing}
F.~R. Hogan, M.~Jenkin, S.~Rezaei-Shoshtari, Y.~Girdhar, D.~Meger, and G.~Dudek, ``Seeing through your skin: Recognizing objects with a novel visuotactile sensor,'' in \emph{Proceedings of the IEEE/CVF Winter Conference on Applications of Computer Vision}, 2021, pp. 1218--1227.

\bibitem{athar2023vistac}
S.~Athar, G.~Patel, Z.~Xu, Q.~Qiu, and Y.~She, ``Vistac toward a unified multimodal sensing finger for robotic manipulation,'' \emph{IEEE Sensors Journal}, vol.~23, no.~20, pp. 25\,440--25\,450, 2023.

\bibitem{zhang2023tirgel}
S.~Zhang, Y.~Sun, J.~Shan, Z.~Chen, F.~Sun, Y.~Yang, and B.~Fang, ``Tirgel: A visuo-tactile sensor with total internal reflection mechanism for external observation and contact detection,'' \emph{IEEE Robotics and Automation Letters}, 2023.

\bibitem{10610373}
Z.~Song, R.~Yu, X.~Zhang, K.~W. Sou, S.~Mu, D.~Peng, X.-P. Zhang, and W.~Ding, ``Satac: A thermoluminescence enabled tactile sensor for concurrent perception of temperature, pressure, and shear,'' in \emph{2024 IEEE International Conference on Robotics and Automation (ICRA)}, 2024, pp. 5680--5686.

\bibitem{10682561}
S.~Li, H.~Yu, G.~Pan, H.~Tang, J.~Zhang, L.~Ye, X.-P. Zhang, and W.~Ding, ``M$^{3}$tac: A multispectral multimodal visuotactile sensor with beyond-human sensory capabilities,'' \emph{IEEE Transactions on Robotics}, vol.~40, pp. 4484--4503, 2024.

\end{thebibliography}

\end{document}